\documentclass[11pt,a4paper,oneside]{book}

\usepackage[numbers]{natbib}  
\usepackage{amsmath,amssymb,amsthm,color,enumerate,graphicx,stackrel,setspace,ifthen,url,array,theorem,pifont}
\usepackage{lscape}
\usepackage{placeins}
\usepackage{multirow}
\usepackage{fancyhdr}
\usepackage{emptypage}
\usepackage{graphicx}
\usepackage{array}
\usepackage[table,xcdraw]{xcolor}
\usepackage{booktabs}  
\usepackage{colortbl}  

\usepackage[inner=3cm,outer=3cm,top=2.5cm,bottom=3cm]{geometry}
\usepackage[sc]{titlesec}
\usepackage{float}

\usepackage[titletoc,title]{appendix}

\usepackage{hyperref}
\usepackage{doi}
\begin{document}
\pagestyle{plain}
\frontmatter

\begin{titlepage} 
	\newcommand{\HRule}{\rule{\linewidth}{0.5mm}} 
	
	\center 
	\textsc{\LARGE University College London}\\[1.5cm] 
	
	
	\textsc{\large A thesis submitted in partial fulfilment of the requirements for the degree of Master of Science in Emerging Digital Technologies, University College London}\\[0.5cm] 
	\HRule\\[0.4cm]
	\textsc{\huge Multimodal Quantum Natural Language Processing: A Novel Framework for using Quantum Methods to Analyse Real Data}\\[0.4cm] 
	\HRule\\[1.5cm]


        \textsc{\Large Hala Hawashin}\\ [0.4cm]
        \textsc{\Large Supervised by \textit{Prof. Mehrnoosh Sadrzadeh}}\\ [0.2cm]
        \textsc{\Large Department of Computer Science}\\[0.2cm]
        \textsc{\Large Faculty of Engineering}\\ [0.7cm]
        


	
        {\large\textit{September 9, 2024}}\\[0.4cm] 

       \includegraphics[width=0.2\textwidth]{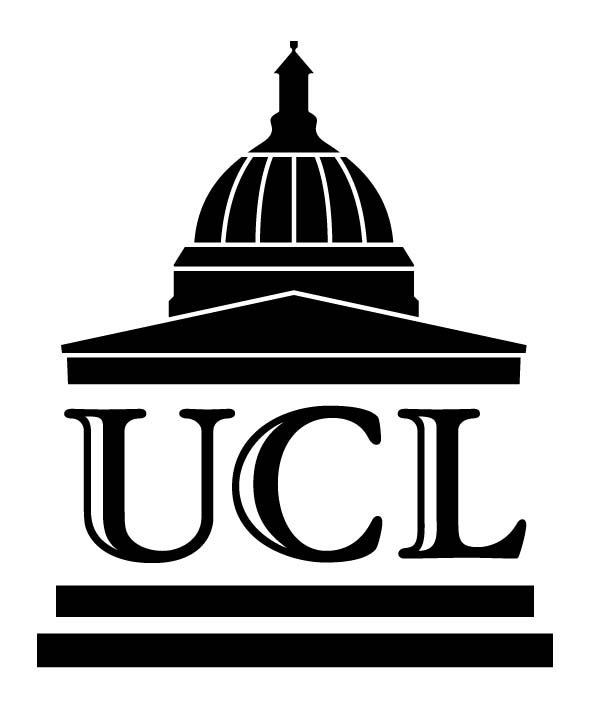}\\[1cm] 
	
	
	 This dissertation is submitted as part requirement for the MSc Emerging Digital Technologies degree at UCL. It is substantially the result of my own work except where explicitly indicated in the text. 
	 

	

	
\end{titlepage}
\clearpage 
\newpage 

 

\section*{Abstract}
Despite significant advances in quantum computing across various domains, research on applying quantum approaches to language compositionality — such as modeling linguistic structures and interactions — remains limited. This gap extends to combining quantum language data with real-world data from sources like images, video and audio. This thesis aims to explore how quantum computational methods can enhance the compositional modeling of language through multimodal data integration. Specifically, we will advance Multimodal Quantum Natural Language Processing (MQNLP) by applying the \texttt{Lambeq} toolkit to image-text classification tasks. We will evaluate syntax-based, bag-of-words, word-sequence, and tree-based models, providing a comparative analysis within a novel MQNLP framework. The experimental design includes two components: Experiment 1, which adopts an unstructured dataset of sentence-image pairs to evaluate models based on verb usage, and Experiment 2, which involves a structured dataset with sentence-image pairs to test models on handling interchangeable subjects and objects.

The results revealed that syntax-based models, specifically DisCoCat and TreeReader, significantly outperformed others by effectively leveraging their understanding of grammatical structure. DisCoCat achieved an average accuracy of 63.18\% on unstructured data, while TreeReader performed well with an average accuracy of 56\% on structured datasets. In contrast, the Bag-of-Words model struggled with structured datasets due to its lack of syntactic awareness, resulting in uniform training metrics across iterations. Sequential models like Cups and Stairs showed limited effectiveness, as their performance did not adequately address the complexities of word composition and structure in both datasets.

Our results suggest that quantum approaches, even with current limitations such as smaller dimensions of image representations, can achieve competitive accuracy compared to classical Multimodal NLP models. This proof of concept highlights the importance of advancing research into multimodal applications within quantum computing, leading to breakthroughs that surpass traditional methods as quantum technology evolves.

\newpage 
\section*{Acknowledgments}

I consider myself incredibly fortunate for the opportunities presented to me in life. All of which would not have been possible without my parents — mama and baba, your support has been the backbone of everything I have achieved and everything I aspire to achieve. A special thank you goes to my cherished friendships, those whose backgrounds may not be scientific but who patiently listened to me discuss this work time and time again.

The guidance of my supervisor, Prof. Mehrnoosh Sadrzadeh, has been instrumental in shaping this thesis, and her support has been truly invaluable. I am also grateful to Dr. Dimitri Kartsaklis, whose informal insightful consultations greatly enriched this work. My deepest thanks go to both for their expertise.

\newpage

\setcounter{tocdepth}{2} 

\cleardoublepage
\tableofcontents

\mainmatter

\chapter{Introduction}
Quantum computing is an emerging technology capable of solving complex problems beyond the reach of even the most powerful classical computers. It involves combining the principles of quantum mechanics with computer science, offering significant impacts in various fields. One such field is Natural Langauge Processing (NLP) which has recently achieved much success through the development of state-of-the-art transformer models, these have led to the development of large language models (LLMs) which are transforming our day to day lives. A persistent challenge in NLP, especially with LLMs, is the “black box” dilemma. This refers to the opaque decision-making processes of these models, which is the result of usage of large and complex deep neural network learning algorithms which have to tune millions of parameters. This has prompted researchers to explore solution to counteract this challenge. A promising solution was presented when researchers discovered that the compositional structure of languages and quantum systems can be modelled within the same mathematical framework, resulting in the creation of \texttt{Lambeq}, a high-level programming toolkit for Quantum Natural Language Processing (QNLP). Although this is advancing the field of NLP in quantum computing, it is also falling behind. The reason for this is that classical NLP research is often combined with other data modalities, including images, video, and audio, leading to the field of Multimodal NLP. In contrast, QNLP has so far been developed primarily for textual data only.

This study aims to advance the field of Multimodal QNLP (MQNLP) by focusing specifically on image-text classification. The goal is to incorporate images alongside text to gain deeper insights using compositional features of language to understand and explain image understanding. Our research introduces a novel framework in MQNLP, utilising the \texttt{Lambeq} toolkit to design multimodal image-text quantum circuits and train their parameters. Another contribution is the design of two multimodal image-text datasets, designed in such a way that — for the first time — allows the QNLP pipeline to process both image and text data. We compare various compositional methods, including syntax-based, bag-of-words, word-sequence, and tree-based models — and analyze their performance on different compositional models for linguistic syntax. Our two datasets led to two experiments, the results of which have been eye-opening. Not only does accounting for the compositional structure of language make the learning process more transparent, but it also results in better prediction accuracy.

The central hypothesis of this study is that syntactic and structure-aware models will outperform models that ignore grammar and sentence composition. These structure-aware models are expected to leverage linguistic patterns to provide more accurate interpretations. Additionally, the study investigates a sub-hypothesis concerning verb composition and understanding. This aspect examines whether the models' ability to differentiate between sentence structures and verb usage significantly enhances their accuracy in interpretation.

\begin{itemize}
 \item \textbf{Experiment 1} uses a dataset containing two images—one positive and one negative—for a given sentence. The sentence structure involves subject-verb-object combinations, where the difference between the two images is the verb used. This approach ensures consistency across the dataset and evaluates the models' ability to learn and distinguish verb-based differences.

 \item \textbf{Experiment 2}  features two sentences—one positive and one negative—paired with a single image. The sentences in this experiment are subject-verb-object combinations, with the subject and object interchanged between the two sentences. This experiment aims to evaluate how the models react when sentence structure and order is critical.
\end{itemize}

The remainder of this study is structured as follows:
\begin{enumerate}
   \item \textit{Literature Review}: This chapter provides a comprehensive overview of the relevant literature, covering the history and evolution of NLP, the basics of quantum computing, and the applications of Multimodal NLP. It aims to establish a solid foundation by exploring existing research and identifying gaps that this study aims to address.
   
    \item \textit{Methodology}: This chapter details the research design and experimental approach, including the creation and utilisation of diagrams that are used to create quantum circuits for their corresponding representation, and the setup for evaluating the models using both unstructured and structured datasets.
    
    \item \textit{Results}: This chapter outlines the details of the experiment, including the justification for the chosen hyperparameters, learning rates, and number of epochs. We also conducted a prediction task and recorded the best-performing and average results for each compositional method across both experiments. We provide tables and diagrams to illustrate the models' convergence and use these visualizations to facilitate a deeper analysis of the outcomes.

    \item \textit{Discussion}: This chapter thoroughly analyzes the performance of various models — syntax-based, tree-based, word-sequence, and bag-of-words—based on experimental results. It also addresses the limitations of the study and identifies areas for improvement and future research.
    
    \item \textit{Conclusion}: The conclusion summarizes the study's key findings and underscore the effectiveness of grammar-aware models in capturing syntactic and semantic relationships for different tasks.

    \item \textit{References}: A list of all sources and references cited throughout the research.
    
    \item \textit{Appendix}: This section includes sample entries of each dataset, the corresponding quantum circuit representations, and a table displaying the results of each model per iteration.

\end{enumerate}


\chapter{Literature Review}
Alan Turing's question, ''Can machines think?'' relates to the field of Natural Language Processing (NLP) as it explores the possibility of machines understanding and interpreting human language \cite{turing1950}. After proposing this question in the 1950s, scientists began their research into the field recognised as artificial intelligence. While the question of whether artificial intelligence can truly be considered intelligent remains a subject of debate, its capacity to revolutionize computing is evident across numerous industries. This literature review focuses on its particular influence within the field of NLP. 


\section{Classical Natural Language Processing}
The evolution of NLP has transformed computers from simple data processors into intelligent systems capable of interpreting, manipulating, and understanding everyday human language \cite{amazon_nlp}. The earliest significant from of NLP was in machine translation, where text was directly translated from Russian to English on a word-for-word basis \cite{Georgetown-IBM2004}. The “mechanical translator” knew only 250 words of Russian and 6 grammar rules to combine them.  At the time, this rule-based system was foundational in the development of machine translation however it was criticised because it required additional manual editing which made the translations slower and more expensive than traditional human translation \cite{hutchinsn.d.}. Government funding was reduced when a report by the Automatic Language Processing Advisory Committee (ALPAC) criticised the project for slow progress \cite{alpac1966}. Consequently, the research directed towards language data understanding diminished until eventually, computers became more powerful and capable of handling more complex tasks. 

In the 1960's, several advancements shaped the field, including ELIZA, one of the first chatbots created to simulate conversation with a human therapist \cite{shrager2024}. Another significant development was SHRDLU, developed by MIT, which used natural language to interact with and manipulate a virtual world of blocks \cite{stanford2024shrdlu}. It was not until the early 2000s where advancements in machine learning prompted researchers to fully explore NLP's potential, enabling the training of larger models and the development of improved architectures which consequently introduced new challenges. 

A common issue faced in training LLMs was that word sequences in test data often differed from those encountered during training due to the exponential growth in the number of possible word combinations. Traditional n-gram models attempted to address this issue by focusing on capturing patterns in short sequences observed during training. Although this approach was effective, the model's capacity was constrained by the size of the dataset, which ultimately limited its performance. This setback, known as the ''curse of dimensionality'', was addressed by Yoshua Bengio et al. in their paper, “A Neural Probabilistic Language Model” \cite{bengio2003neural}. They introduced a transformative approach by leveraging neural networks to simultaneously learn a distributed representation of words and a probability function for word sequences. This method reduced the high-dimensional word representations used in n-gram models, significantly enhancing the model's computational efficiency. It also improved the model's ability to generalize to unseen word sequences, thereby advancing the state-of-the-art beyond n-gram models.

This laid the groundwork for research into capturing the distributional characteristics of words and measuring their similarity. A paper published by Google introduced the skip-gram model, which similarly used a neural network architecture to learn high-quality word representations from unstructured data \cite{google2013} . This technique represents words in a continuous vector space, where the similarities between words can be captured through vector calculations. For example, vector operations like vec(''Madrid'') - vec(''Spain'') + vec(''France'') result in a vector closer to vec(''Paris'') than to any other word vector. The promising results, which included increased accuracy and faster computation times, were further developed in the paper ''Distributed Representations of Words and Phrases and their Compositionality'' in 2013 which led to the creation of the \texttt{Word2vec} toolkit, this significantly raised interest in word embedding among researchers and enabled practical applications \cite{google2013a}. 

In 2014, Seq2Seq models were introduced, primarily for tasks such as machine translation (\cite{bahdanau2014neural} \cite{sutskever2014sequence}) and text summarizing (\cite{see2017get}, \cite{rush2015neural}). These models were based on recurrent neural networks (RNNs), which facilitate the transfer of information from one step of a sequence to the next using encoder and decoder architectures. At each time step, the current input is processed along with the hidden state from the previous step to produce an output and update the hidden state, allowing the model to maintain a memory of previous inputs. 

Earlier RNNs had limited memory and often struggle with retaining information over long sequences, a modified version was introduced by \cite{gers1999learning} adding a Long Short-Term Memory (LSTM) network to capture dependencies over much longer timelines. The Seq2Seq models leveraged LSTM networks for both the encoder and decoder, thereby enhancing their ability to handle long-range dependencies and slightly improving their performance. Despite these improvements introduced by LSTMs, there remained a small but notable issue affecting model performance. Specifically, the compression of the entire input sequence into a fixed-size vector representation could lead to the loss of important information in long sequences, impacting the model's ability to fully retain and utilise context \cite{bui2021multi}.

\begin{figure}[H] 
    \centering 
    \includegraphics[width=0.5\textwidth]{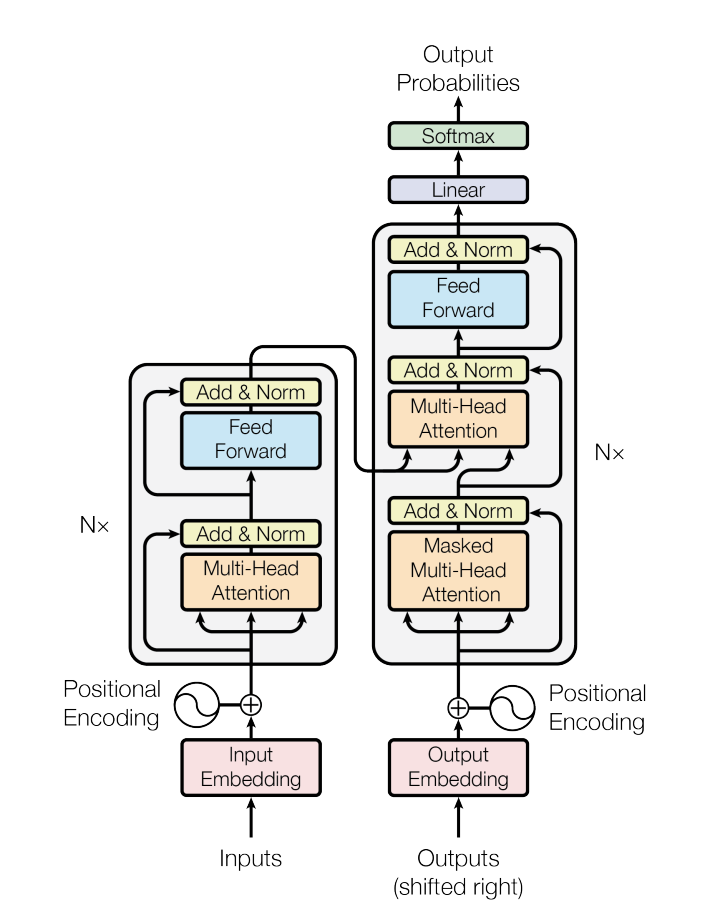} 
    \caption{This diagram shows the architecture of the Transformer model as described in the paper "\textit{Attention is All You Need}" \cite{attention2017}.} 
    \label{fig:transformer} 
\end{figure}

In 2015, Vaswani et al. introduced the Transformer model, which has since become the backbone of many state-of-the-art NLP systems due to its ability to enable parallel processing of input data \cite{attention2017}. The Transformer model is built on a self-attention mechanism, which allows the model to process input data more efficiently by enabling each position in the input sequence to attend to all other positions. This mechanism facilitates the capture of dependencies regardless of their distance within the sequence. The Transformer architecture also comprises an encoder and a decoder, each consisting of multiple layers that include multi-head self-attention and feed-forward neural networks as shown in Figure \ref{fig:transformer} \cite{whites2020benefit}. In 2018, Google introduced BERT (Bidirectional Encoder Representations from Transformers), which pioneered bidirectional training of transformers and greatly advanced the adoption of pre-trained models \cite{devlin2018bert}. BERT's approach allowed for a deeper understanding of context by considering both the left and right context of a word simultaneously. Following this, in 2019, OpenAI released GPT-2 \cite{gpt-2}, and later GPT-3 \cite{gpt-3}, large-scale generative models capable of producing coherent and contextually relevant text across a variety of tasks. These advancements marked a significant leap in the development of state-of-the-art language models, influencing various applications and driving further research and adaptation in the field.

\section{Multimodal Learning in Natural Language Processing}
Naturally, advancements in the linguistic understanding of natural language paved the way for exploring the benefits of incorporating images to provide additional context to textual data. This motivated researchers to adapt techniques previously developed for vision learning and apply them to multimodal research.


\subsection{Vision Learning} 
\label{visionlearning}
The earliest research in computer vision dates back to the early development of artificial neural networks by Frank Rosenblatt \cite{ann1958} who drew inspiration from psychological theories of perception. Rosenblatt's work was based on the idea of simulating the brain's thought processes and how it recognizes patterns, which laid the foundation for the Perceptron model, a fundamental building block for early vision learning systems. Similar to advancements in NLPs, progress in computer vision was initially constrained by computational power. 

In the 1970s, despite advances in computer technology, limitations on data storage and computing powers persisted. This led researchers to focus on detecting and extracting relevant features from images, resulting in pioneering work in early feature extraction mechanisms, specifically corner and edge detection \cite{roberts1963, prewitt1970, sobel1970, canny1986, moravec1980, harris1988, marr1980, marr1982}. Subsequent years saw significant progress in the field, with a shift towards more complex tasks such as object tracking, driven by advancements in machine learning algorithms \cite{shi1994good}, \cite{yilmaz2006object}, \cite{viola2001rapid}.

At its core, the power of vision learning lies in its ability to capture and interpret crucial features from visual data. A significant advancement in the underlying architecture of successful models is attributed to Convolutional Neural Networks (CNNs). The pivotal role of CNNs in visual recognition tasks was first highlighted in the 1990s with the introduction of LeNet-5, a CNN model specifically designed for handwritten digit recognition. LeNet-5 showcased how CNNs utilise convolutional layers to apply filters to input data, generating feature maps that reveal various aspects and details of an image. Five convolutions layers used were followed by pooling layers, which reduce the dimensionality of the data and enhance the network's efficiency. This foundational work, was detailed in the paper ''Handwritten Digit Recognition with a Back-Propagation Network'' by \cite{lecun1989handwritten} which highlighted the use of backpropagation for training neural networks. The primary application of the novel CNN model was recognizing handwritten digits from the MNIST dataset, which have shown a high accuracy and a 1\% error rate.

\begin{figure}
    \centering
    \includegraphics[width=0.8\linewidth]{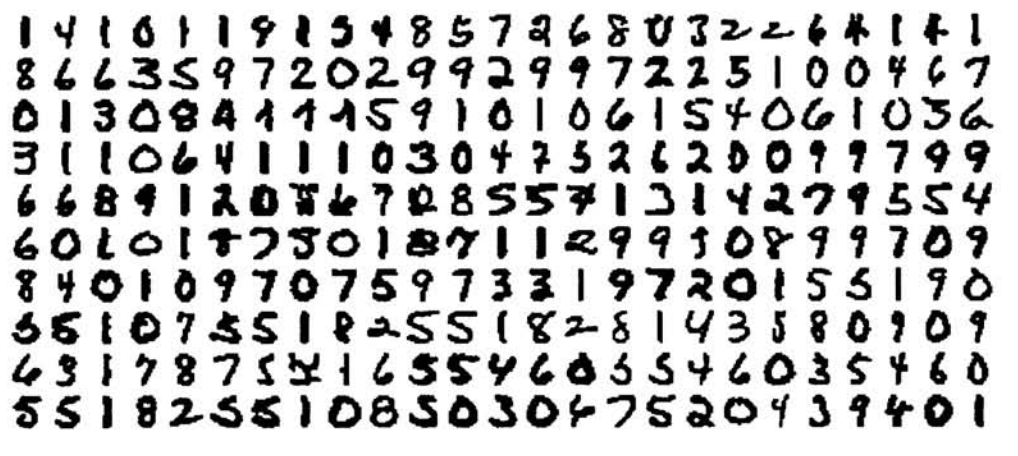}
    \caption{This figure illustrates an example of the test set used in the paper "Handwritten Digit Recognition with a Back-Propagation Network" \cite{lecun1989handwritten}. }
    \label{fig:enter-label}
\end{figure}

Building on this foundation, researchers at the University of Toronto developed AlexNet by conducting an extensive image classification study with over a million images \cite{krizhevsky2012imagenet}. The model showed the effectiveness of deep neural networks by using Rectified Linear Unit (ReLU) activation's and dropout for regularization. These techniques helped reduce overfitting and improve the model's performance through the use of five convolutional layers, followed by max-pooling layers, and three fully connected layers. Its architecture allowed it to excel in large-scale image classification tasks, ultimately securing first place at the ImageNet Large Scale Visual Recognition Challenge (ILSVRC) in 2012 \cite{medium_alexnet_review}. Despite its groundbreaking impact, AlexNet faced limitations due to its relatively shallow depth and its struggle to capture very fine details. These shortcomings prompted the development of the VGGNet architecture two years later.

\begin{figure}
    \centering
    \includegraphics[width=1.0\linewidth]{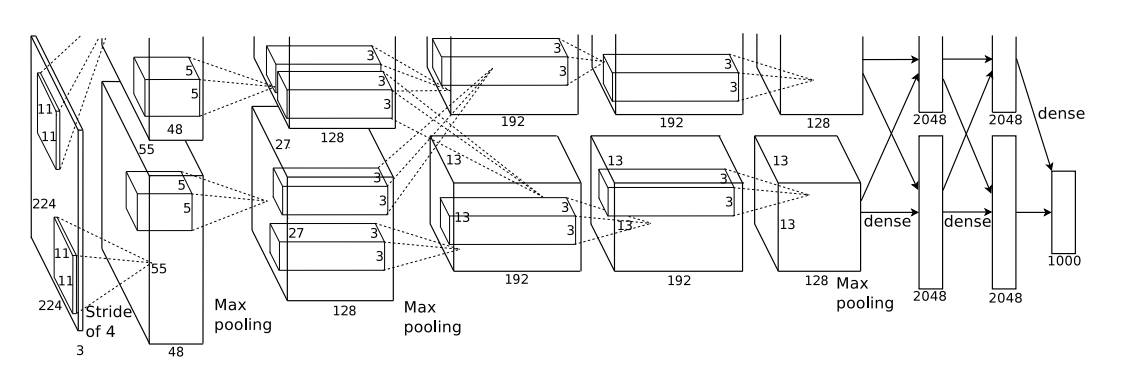}
    \caption{This diagram shows the architecture of AlexNet, as introduced in the paper by \cite{krizhevsky2012imagenet} for the ImageNet competition.}
    \label{fig:AlexNet}
\end{figure}

Introduced in 2014 by \cite{simonyan2014very}, VGGNet increased the depth of the CNN while maintaining a uniform architecture. VGGNet utilised very small (3x3) convolutional filters and added more layers, resulting in a deeper and more detailed network. This depth allowed VGGNet to capture more intricate patterns and details in images, thereby outperforming AlexNet, specifically in object recognition tasks. The increased depth, however, also brought substantial computational and memory costs. The large number of parameters made VGGNet limited in its practical application in environments with constrained resources.

\begin{figure}
    \centering
    \includegraphics[width=0.6\linewidth]{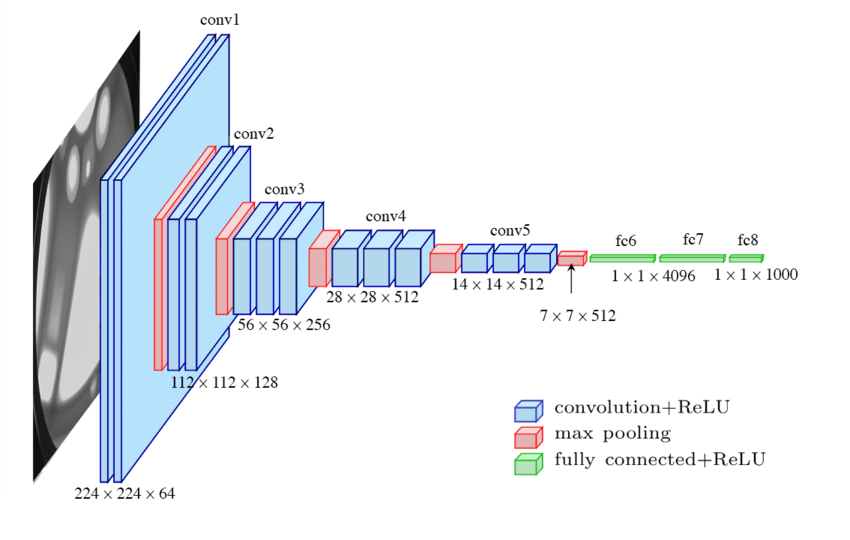}
    \caption{This diagrams represents the architecture of VGGNet as explained in the article by \cite{VGGNet}.}
    \label{fig:VGGNet}
\end{figure}

\begin{figure}
    \centering
    \includegraphics[width=1.0\linewidth]{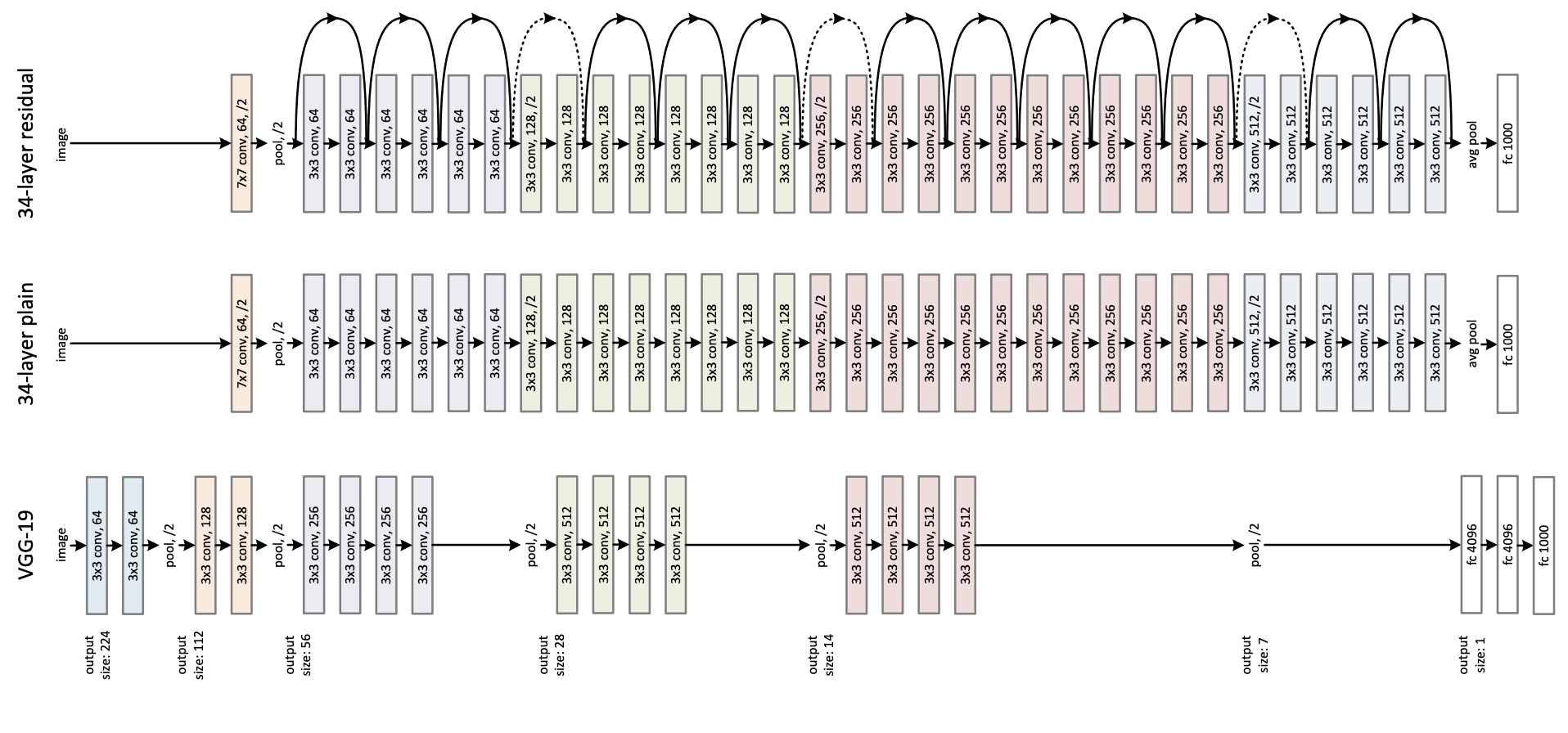}
    \caption{This figure shows the ResNet-15 architecture, as presented in \cite{ResNet2016}.}
    \label{fig:ResNet}
\end{figure}

One year later, \cite{ResNet2016} made a significant advancement by addressing the vanishing gradient problem and other limitations of VGGNet through the introduction of skip connections or residual blocks. These residual blocks are a key feature of ResNet that allow the network to bypass one or more layers during the forward pass by adding the input of a layer directly to its output as shown in Figure \ref{fig:ResNet}. This enabled the training of much deeper models without encountering the vanishing gradient issue, leading to substantial improvements in performance on complex tasks and resulting in the ResNet architecture. Although this architecture still required careful tuning to balance depth and complexity, it remains one of the leading architectures and continues to be used effectively for various image recognition and classification tasks today.

\subsection{Applications of Multimodal Natural Language Processisng}

As evident from the advancements in both fields, NLPs and computer vision tasks have progressed at a similar pace. The current state of both have allowed researchers to leverage insights from textual and visual data into multi modal learning approaches. This section will address the most common application: Visual Question Answering (VQA), image captioning, object detection, and sentiment analysis.

A popular area of research within multimodal learning is VQA and image captioning, where models are trained to understand and describe the content of images using natural language. For example, the 2019 study 'Unified Vision-Language Pre-Training for Image Captioning and VQA' introduces a unified framework for pre-training models that can be fine-tuned for both tasks \cite{zhou2019unified}. This approach leverages a shared transformer-based architecture that learns to understand and generate language from visual inputs. In practical applications, such models are used in various domains, such as improving the generation of descriptive captions for visually impaired users \cite{Chen2022} and enhancing consumer experiences and extract relevant feedback segments \cite{amazon2022insight}. 

In some cases, models must effectively handle data they have never encountered before, a challenge known as zero-shot learning. This capability is crucial for object detection, particularly in the medical field, where models must identify and analyze new types of medical images or conditions that were not part of the training data. A notable advancement in this field is the 2022 paper on Contrastive Language–Image Pretraining (CLIP) \cite{CLIP2021}. CLIP revolutionizes vision tasks by training models to connect images with textual descriptions, allowing for zero-shot learning across various tasks without needing specific training data for each task. The results of CLIP demonstrated its versatility, as it performed competitively across multiple vision benchmarks, including object detection. Its ability to generalize effectively to new tasks with minimal additional training has made it a landmark achievement, especially in the medical field for detecting cancerous cells \cite{cancer-2017detecting}; \cite{cancer-2019attention}.

Multimodal sentiment analysis has gained significant traction due to the increasing need to automate the assessment of user sentiment toward products or services. A notable example of this is the paper ''Multimodal Sentiment Analysis Using Hierarchical Fusion with Context Modeling'' \cite{li2022multimodal}, which explores advanced methods for integrating various data sources to improve sentiment analysis. This work presents a model that combines textual, visual, and acoustic information for sentiment analysis. It has been applied in real-world scenarios such as analyzing sentiment in social media videos, including YouTube vlogs and product reviews, where it combines the speaker's facial expressions, tone of voice, and spoken content to deliver a more accurate sentiment analysis than text-only methods \cite{Sentiment2022}. 


\section{Quantum Inspired Approach}

The advancements in NLP and multimodal NLP within classical applications have inspired researchers to explore quantum-inspired approaches. Although quantum computing is still in its early stages and currently does not match the processing power of classical systems, it has shown significant potential through promising results. This suggests that, in the near-future, quantum computing could eventually surpass classical methods, offering transformative possibilities for the field.


\subsection{Basics of Quantum Computing}
The origin of quantum computing can be traced back to the early work of David Hilbert, a German mathematician, who in the early 20th century defined and developed the concept of a Hilbert space \cite{hilbert1932}. Hilbert's work on infinite-dimensional spaces was groundbreaking, as it introduced key mathematical concepts such as orthogonality, completeness, and inner products that would prove essential in the mathematical formulation of quantum theory. This concept was later refined and generalized by John von Neumann, who is also credited with developing the architecture that forms the basis of modern-day computers. In his book ''Mathematical Foundations of Quantum Mechanics'', von Neumann provided the mathematical framework for representing quantum systems using a Hilbert space, laying the groundwork for quantum mechanics \cite{vonNeumann1932}.

A Hilbert space, also known as a state space, is a complete, infinite-dimensional vector space equipped with an inner product, which allows for the calculation of lengths (norms) and angles (inner products) between vectors within the space \cite{Griffiths2024, Halmos1951}. This concept is foundational to quantum computing, where the qubit—the basic unit of quantum information—is represented as a vector in a two-dimensional complex state space. In quantum mechanics, this space is used to describe the quantum state of a spin, angular momentum of particles, providing the physical and mathematical basis for \textit{qubits} in quantum computing \cite{Griffiths2024}. 

Unlike the conventional binary operators used in traditional computers, quantum information utilises standard basis vectors denoted by $\vert 0 \rangle$ and $\vert 1 \rangle$. This notation, known as dirac or bra-ket notation, was introduced by \cite{Dirac1930} and is widely used as the standard mathematical notation in quantum mechanics to represent quantum states . These qubits enable quantum computers to perform computations that are infeasible for classical computers by utilising unique features of quantum mechanics, such as superposition and entanglement.  

Superposition allows a qubit to exist in a linear combination of both $\vert 0 \rangle$ and $\vert 1 \rangle$ states simultaneously, rather than being limited to a single binary state. This allows quantum algorithms to explore many possible solutions or paths at once. For a single qubit, this superposition can be represented as:
\[
|\psi\rangle = \alpha |0\rangle + \beta |1\rangle
\]
where \(|0\rangle\) and \(|1\rangle\) are the basis states of the qubit, and \(\alpha\) and \(\beta\) are complex coefficients such that:
\[
|\alpha|^2 + |\beta|^2 = 1
\]

Entanglement refers to the phenomenon where qubits become intertwined in such a way that the state of one qubit instantly influences the state of another, regardless of the distance between them. This allows for faster information processing and communication, forming the foundation for many quantum algorithms that outperform their classical counterparts \cite{Nielsen2010, Dirac1930}.

Qubits are manipulated through quantum circuits to perform complex computations. The first experimental realization of a quantum circuit was conducted by researchers at IBM and Stanford University in 1998 \cite{IBM2021}. This experiment involved three key components: \textit{qubits}, which represent quantum states in the system; \textit{quantum gates}, which are operations applied to the qubits and depicted by symbols on circuit diagrams; and\textit{ measurements}, which determine the final states of the qubits to obtain classical bits \cite{Nielsen2000}. Similar to classical logic gates like AND, OR, and NOT, quantum gates perform linear operations as unitary operators on qubits. Examples of quantum gates include the Pauli gates (X, Y, Z), the Hadamard gate, and the CNOT gate \cite{Gates2020}. 

The Pauli-X gate, commonly referred to as the quantum NOT gate, flips the state of the qubit similar to the classical NOT gate. If the qubit is in the state $\vert 0 \rangle$, applying the Pauli-X gate changes it to $\vert 1 \rangle$ and vice versa. The matrix representation of the Pauli-X gate is:
\[
X = 
\begin{pmatrix}
0 & 1 \\
1 & 0
\end{pmatrix}
\]

The Pauli-Y gate rotates the qubit's state by 90 degrees around the Y-axis of the Bloch sphere, and adds a phase shift. The matrix representation of the Pauli-Y gate is:
\[
Y = 
\begin{pmatrix}
0 & -i \\
i & 0
\end{pmatrix}
\]

The Pauli-Z gate acts by introducing a phase shift of $\pi$ to the $\vert 1 \rangle$ state, effectively flipping the sign of its amplitude. Otherwise, the $\vert 0 \rangle$ state remains unchanged. The matrix representation of the Pauli-Z gate is:
\[
Z = 
\begin{pmatrix}
1 & 0 \\
0 & -1
\end{pmatrix}
\]

The Hadamard gate creates an equal superposition of the $\vert 0 \rangle$ and $\vert 1 \rangle$ states. Specifically, it transforms the basis states $\vert 0 \rangle$ and $\vert 1 \rangle$ into the superposition states $\vert + \rangle$ and $\vert - \rangle$, respectively. The matrix representation of the Hadamard gate is:
\[
H = \frac{1}{\sqrt{2}} 
\begin{pmatrix}
1 & 1 \\
1 & -1
\end{pmatrix}
\]

Controlled gates involve two qubits: a control qubit and a target qubit. The operation on the target qubit depends on the state of the control qubit. Specifically, if the control qubit is in the state $\vert 1 \rangle$, the Controlled-NOT (CNOT) gate flips the target qubit, otherwise - if the control qubit is in the state $\vert 0 \rangle$ - it remains unchanged. The matrix representation of the Controlled-NOT gate is:
\[
\text{CNOT} = 
\begin{pmatrix}
1 & 0 & 0 & 0 \\
0 & 1 & 0 & 0 \\
0 & 0 & 0 & 1 \\
0 & 0 & 1 & 0
\end{pmatrix}
\]

A collection of quantum gates arranged in a specific sequence forms a quantum circuit. This circuit can create and manipulate superpositions of qubit states by starting with qubits in an initial state known as an \textit{ansatz}. In this context, an \textit{ansatz} refers to a parameterized quantum state or a specific form of quantum circuit used as an initial approximation in a variational quantum algorithm \cite{PennyLane2024}. The quantum circuit then applies a series of quantum gates to evolve the qubits, followed by measuring the output qubits to obtain the final result.

Two of the most significant applications of quantum algorithms that leverage these principles are Shor's algorithm and Grover's algorithm. Shor's algorithm is designed to factorize large integers into their prime factors by exploiting quantum superposition to simultaneously perform computations on multiple possibilities \cite{shor1997}. Unlike classical algorithms that require exponential time to factorize large integers, Shor's algorithm can accomplish this in logarithmic computation time, specifically in $O(\log^3 N)$ time and $O(\log N)$ space. This significant reduction in computation time challenges is widely used in cryptographic schemes like RSA, which relies on the difficulty of factoring large numbers and is employed by corporations and banks for secure communications.

Similarly, Grover's algorithm provides a quadratic speedup for unstructured search problems by utilizing superposition to search an unsorted database in. In his 1996 paper, “\textit{A Fast Quantum Mechanical Algorithm for Database Search}", \cite{grover1996} demonstrated that his algorithm provides a quadratic speedup for unstructured search problems \(O(N \sqrt{N})\) time, where \textit{N} is the number of items. This efficiency has been applied to searching through linguistic databases, such as locating specific patterns or keywords in large text corpora, demonstrating the potential for quantum algorithms to advance computational capabilities in NLP.

This breakthrough in quantum computing not only demonstrated the transformative potential of quantum algorithms but also inspired researchers to explore their applications in other fields, including NLP.

\subsection{Quantum Natural Language Processing}

QNLP represents a novel intersection of quantum computing and NLP, where principles from quantum mechanics are applied to enhance the modeling and interpretation of language. 

A pivotal work in this area is \textit{“Mathematical Foundations for a Compositional Distributional Model of Meaning”,} which introduces a method for employing quantum circuits to model the compositional structure of language \cite{mathfoundation2010}. This method effectively captures how the meanings of individual words combine to form the meaning of a sentence. It aligns with categorical models of meaning in NLP, where grammar is understood as a compositional structure, providing a more structured and interpretable approach to language processing. The theoretical foundations of this is deeply embedded in category theory, particularly its applications to linguistics and computation.

Category theory, a branch of mathematics focused on abstract structures and their relationships, provides the foundation for this approach. It uses two key components: \textbf{objects}, representing mathematical structures (e.g., sets, spaces, or linguistic units like words and sentences in QNLP), and \textbf{morphisms}, or arrows, which describe the relationships or transformations between objects, such as grammatical or syntactic connections in language. This is further explained in section \ref{CategoryTheory}.

A key concept within this framework is the monoidal category, which extends category theory to encompass tensor products, thereby providing a structure to model interactions between objects. In QNLP, monoidal categories play a crucial role in representing linguistic compositionality, enabling the combination of word meanings into coherent sentence-level interpretations. This approach is grounded in pregroup grammar, an algebraic structure introduced by \cite{lambek1999}, which is fundamental for modeling syntactic structures. \cite{bobcoecke2010categorical} highlighted a significant challenge in NLP: while dictionaries exist for individual words, there is a lack of similar resources for entire sentences. This gap is effectively addressed by Lambek’s work which assigns types to words and uses reductions to verify grammatical correctness, forming the backbone of syntactic analysis in the Distributional Compositional Categorical (DisCoCat) model.

The DisCoCat model integrates distributional semantics with category theory, providing a rigorous mathematical framework for modeling meaning in language. This approach addresses the need for more interpretable models in NLP, as traditional NLP models, often based on deep neural networks, are frequently criticized for being "black boxes" due to their lack of interpretability. DisCoCat enhances transparency by explicitly incorporating linguistic structures, such as grammar and syntax, into distributional language models. 

The earliest application of DisCoCat on a Noisy Intermediate-Scale Quantum (NISQ) processor was demonstrated in the paper “Grammar-aware Sentence Classification on Quantum Computers” \cite{grammareAware}, using IBM's quantum hardware. This study encoded sentences as parameterized quantum circuits and used entangling operations to model grammatical structures. The results demonstrated that quantum circuits could handle NLP tasks with expressiveness and precision comparable to, and sometimes surpassing, classical methods. The \texttt{Lambeq} toolkit, as discussed by \cite{lambeq2021}, has been utilised in quantum machine learning to capture semantic meaning more effectively than classical methods, which can be found in the work of \cite{QML}.

Overall, \texttt{Lambeq} has been pivotal in demonstrating the use of quantum technology in NLP tasks, showcasing how quantum computing can transform NLP through enhanced performance. The toolkit has been designed to implement a range of models—including syntax-based, bag-of-words, and sequence models—on both quantum simulators and computers \cite{lambeq_tutorial}. By providing a practical foundation for QNLP research, \texttt{Lambeq} has enabled significant progress in the field, facilitating the integration of quantum computing techniques into NLP.

\subsection{Current Research on Multimodal Quantum Applications}
Building on this foundation, there has been some recent research in quantum image processing which led to the development of various novel techniques. A study on quantum image translation (QIT) introduced the concept of translating quantum images using specialized circuits \cite{quantum_image_translation}. Another development is in quantum image encryption and decryption for cryptography, where algorithms leveraging quantum geometric transformations were proposed to enhance data security and confidentiality by encoding image properties in quantum states \cite{quantum_image_encryption}. Additionally, early research on quantum generative adversarial networks (GANs) has demonstrated their ability to generate high-dimensional images, with quantum GANs performing competitively against classical GANs in image generation tasks, thereby highlighting the practical potential of quantum machine learning \cite{QGeneartive2021}.

As for multimodal approaches most research is directed towards sentiment analysis tasks, demonstrating promising results. For instance, the ''Quantum-Like Multimodal Network (QMN)'' framework employs density matrix-based CNNs and LSTM networks to integrate and analyze text and image data, achieving a 7\% improvement in classification accuracy and an 8\% boost in F1-score compared to traditional models \cite{zhang2020quantum}. Similarly, the Quantum-Enhanced Multi-Modal Analysis (QeMMA) framework utilises a Variational Quantum Circuit (VQC) combined with a recurrent neural network to capture complex multi-modal interactions, surpassing baseline models with a 3.5\% increase in accuracy and a 10\% improvement in F1-score \cite{phukan2023qemma}. Additionally, the Quantum Multi-Modal Data Fusion (QuMIN) model showed significant advancements in humor detection, improving precision by 12\% and F1-score by 11\%  \cite{phukan2024qumin}. These studies collectively highlight the potential of quantum-inspired methodologies in enhancing multi-modal sentiment analysis and addressing intricate linguistic interactions. 
\chapter{Methodology}
The following section outlines the methodology of the MQNLP study and provides a detailed description of our approach. We use a diagrammatic method that integrates two distinct sub-diagrams: one representing textual information and another for visual information. These sub-diagrams are combined through a merging box, which then undergoes transformation into a parameterized quantum circuit representation for training. We will train these diagrams using four different compositional models to conduct a comprehensive analysis of their performance.


\section{Diagram Representation}
Diagrams are a powerful graphical tool derived from category theory, crucial for visualizing complex relationships in QNLP. This approach simplifies the understanding of intricate structures by depicting them through intuitive visual elements. The \texttt{Lambeq} toolkit was mainly designed to work with linguistic data types but is also adaptable to arbitrary atomic types, which can be used to represent image data.

\subsection{Category Theory}
\label{CategoryTheory}
At its core, category theory has three fundamental components: objects, morphisms, and composition.

\begin{itemize}
\item \textbf{Objects} in category theory can be thought of as types, entities, or data points. They serve as the foundational elements that the theory operates on. For example, in the context of QNLP, objects might represent linguistic elements such as words or phrases. In a string diagram, these objects are typically depicted as points, lines, or other shapes, providing a visual representation for the relationships being modeled. 
\item \textbf{Morphisms}, on the other hand, represent the processes or functions that map one object to another. They are the "actions" that connect objects, showing how one type of entity is transformed into another within the framework. In QNLP, morphisms could represent the grammatical rules or semantic relationships that link different linguistic elements. In a string diagram, morphisms are illustrated as arrows or boxes that connect the objects, visually indicating the direction and nature of the transformation.
\item \textbf{Composition} involves applying one morphism after another, essentially chaining together a series of transformations. For example, if we have a morphism $f: A \rightarrow B$ and another morphism $g: B \rightarrow C$, their composition $g \circ f: A \rightarrow C$ can be represented in a string diagram by drawing an arrow from $A$ to $B$, followed by another from $B$ to $C$, with the arrows meeting at $B$. 
\end{itemize}

In many ways, category theory aligns closely with the compositional nature of language. To illustrate this, string diagrams have been developed to depict and facilitate the reading of categorical computations. These diagrams represent how words interact and combine to form sentences, similar to how morphisms map objects in category theory. This approach also mirrors quantum circuits, where quantum gates manipulate qubits to perform computations. Thereby allowing the compositional nature of language to be modeled like quantum computation, with operations (morphisms) applied to quantum states (objects) in a structured, sequential manner.

A string diagram primarily consists of boxes and lines. \textit{Boxes} represent individual entities or operations within the diagram, such as words or grammatical functions. Each box corresponds to a specific component of the linguistic structure, like a noun or verb. \textit{Lines} illustrate the flow of information or meaning between these boxes, showing how different words or elements are connected within a sentence or grammatical structure.

To represent specific types of interactions or relationships between objects, \textit{cups} and \textit{caps} are used. A \textit{cup} is a curved structure resembling a "U" and signifies a particular type of connection between two objects, illustrating how they combine according to grammatical rules. Conversely, a \textit{cap} is oriented in the opposite direction, resembling an inverted "U." \textit{Caps} represent dual relationships or interactions, often indicating how elements are separated or diverged.

To illustrate the concepts above with an example, consider the sentence \texttt{Dog chases cat}. Here, \texttt{Dog} and \texttt{cat} are objects of type \( N \) (noun), and \texttt{chases} is a morphism connecting these nouns. In this context, \( n^r \) and \( n^l \) represent the specific roles of these nouns relative to the verb \texttt{chases}. Specifically, \( n^r \) indicates the noun on the right (\texttt{cat}), which is the object of the verb, while \( n^l \) represents the noun on the left (\texttt{Dog}), which is the subject performing the action. The diagram would show an arrow from \texttt{Dog} to \texttt{chases} (indicating \( n^l \)) and another from \texttt{chases} to \texttt{cat} (indicating \( n^r \)), visually representing how the morphism \texttt{chases} combines the noun objects to form a coherent sentence. This visual representation simplifies understanding by breaking down the sentence structure into its component parts and their relationships.

\begin{figure}[h!] 
    \centering 
    \includegraphics[width=1.0\textwidth]{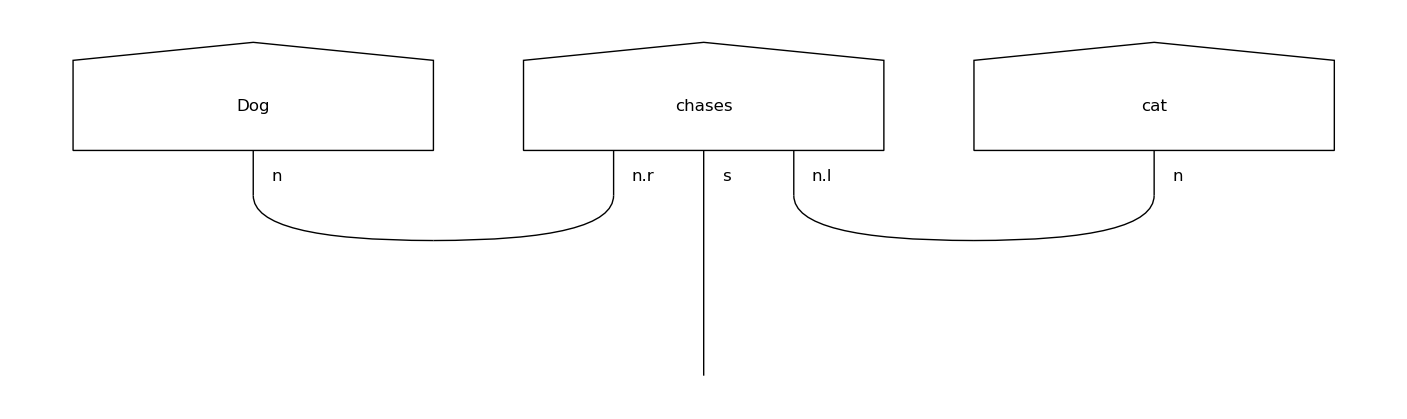} 
    \caption{This diagram illustrates the DisCoCat string diagram representation for the sentence "Dog chases cat".} 
    \label{fig:dog-chase-cat} 
\end{figure}


\subsection{Compositional Models}

The \texttt{lambeq} documentation, as discussed in \cite{QNLPinPractice}, highlights compositional models that can be applied, each with different syntactic information. \cite{lambeq_tutorial}. 

\textbf{Syntax-based models} focus on the grammatical structure of sentences, utilizing grammatical rules that govern the arrangement of words, phrases, and clauses to construct meaningful sentences. \texttt{Lambeq} deploys a Distributive Compositional Categorical (DisCoCat) model which integrates distributional semantics with categorical compositionality to model the meaning of natural language. It employs functors to map diagrams from the category of pregroup grammars to vector space semantics. This process is facilitated by the \texttt{BobcatParser} class, which generates a Combinatory Categorial Grammar (CCG) derivation for a sentence and converts it into a string diagram. The CCG deriavtion operates by assigning categories to words and then applying combinatory rules to determine how these categories combine to form larger syntactic units. This method allows for a detailed and structured representation of grammatical relationships within sentences. The diagrammatic representation is shown in Figure \ref{fig:dog-chase-cat}. 

\begin{figure}[h!] 
    \centering 
    \includegraphics[width=0.8\textwidth]{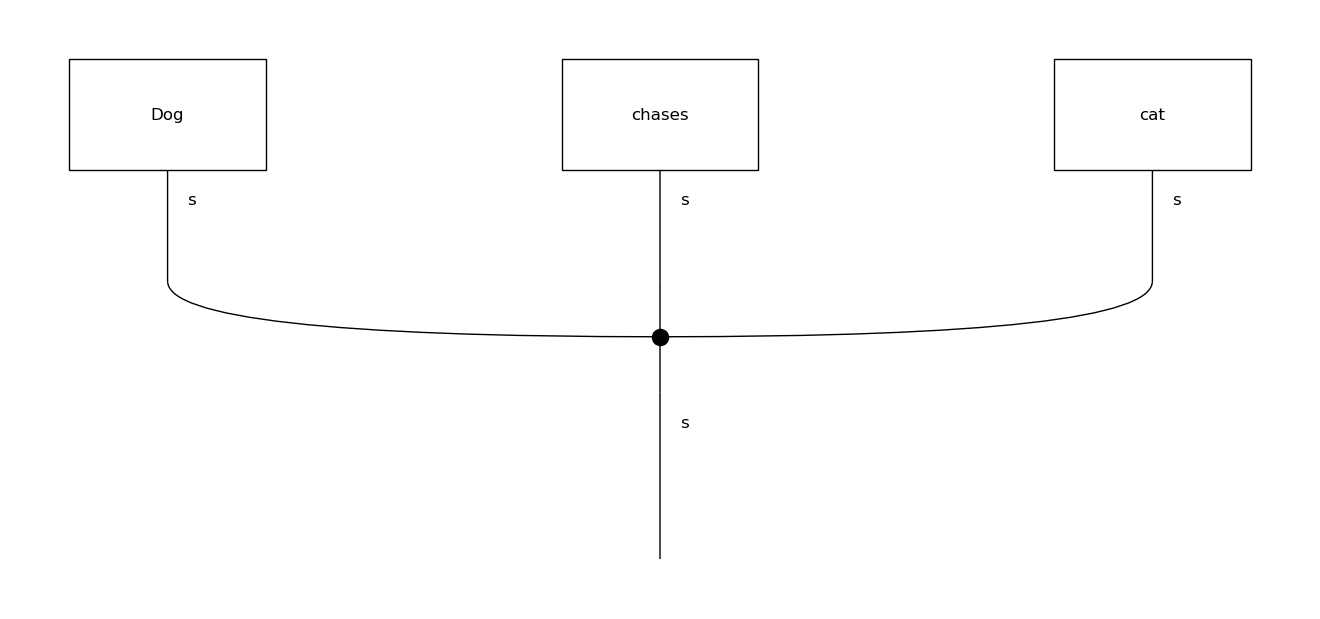} 
    \caption{This diagram shows the string diagram representation of the Spider model for the sentence ``Dog chases cat''.} 
    \label{fig:spider-model} 
\end{figure}

The\textbf{ Bag-of-Words model} represents text as a collection of individual words without considering their syntactic or semantic relationships. This model is particularly useful for tasks such as text classification where the focus is solely on the presence of words, making it less effective for understanding grammatical roles and sentence structure. In the \texttt{Lambeq} toolkit, the \textit{Spider model} implements the bag-of-words approach, visualizing text in a diagrammatic format suitable for quantum circuit transformations, shown in Figure \ref{fig:spider-model}.

\begin{figure}[h!] 
    \centering 
    \includegraphics[width=0.8\textwidth]{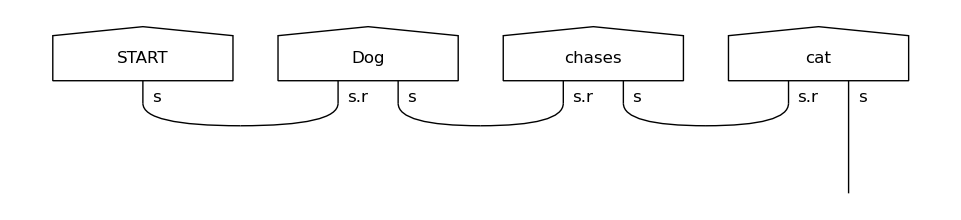} 
    \caption{This diagram shows that string diagram representation of the Cups model for the sentence ``Dog chases cat''.} 
    \label{fig:cups-model} 
\end{figure}

\begin{figure}[h!] 
    \centering 
    \includegraphics[width=0.8\textwidth]{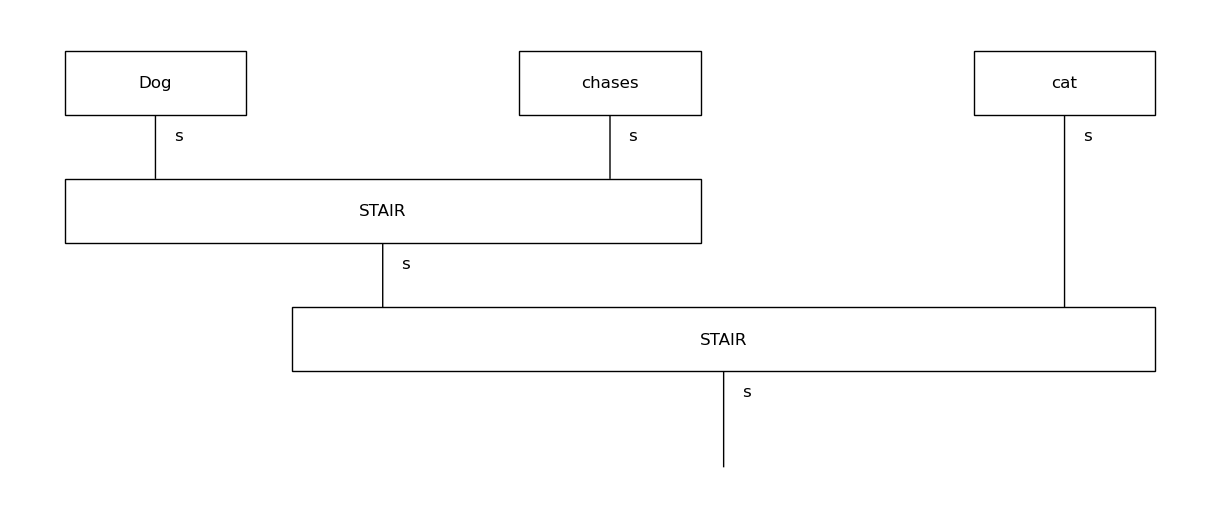} 
    \caption{This diagram shows that string diagram representation of the Stairs model for the sentence ``Dog chases cat''.} 
    \label{fig:stairs-model} 
\end{figure}

\newpage
\textbf{Word-sequence models} address the order and relationships of words within sentences in two different ways. The \textit{Cups Reader} model emphasizes the immediate sequence of words, analyzing consecutive word pairs to understand how the order affects meaning. The \textit{Stairs Reader} model uses a hierarchical structure to break down text into layers or ``steps,'' capturing complex semantic relationships (refer to Figure \ref{fig:cups-model}, \ref{fig:stairs-model}). 

\newpage
The \textbf{Tree-based model} also leverages the structured nature of CCG derivations, similar to the syntax-based model. However, these derivations are organized to allow for a direct interpretation of compositional steps without requiring additional transformations, such as those used in syntax-based models with pre-group grammar, which represents syntactic rules through objects and morphisms. This approach simplifies the process by avoiding extra steps, making the interpretation and application of compositional models more straightforward.

\begin{figure}[h!] 
    \centering 
    \includegraphics[width=0.8\textwidth]{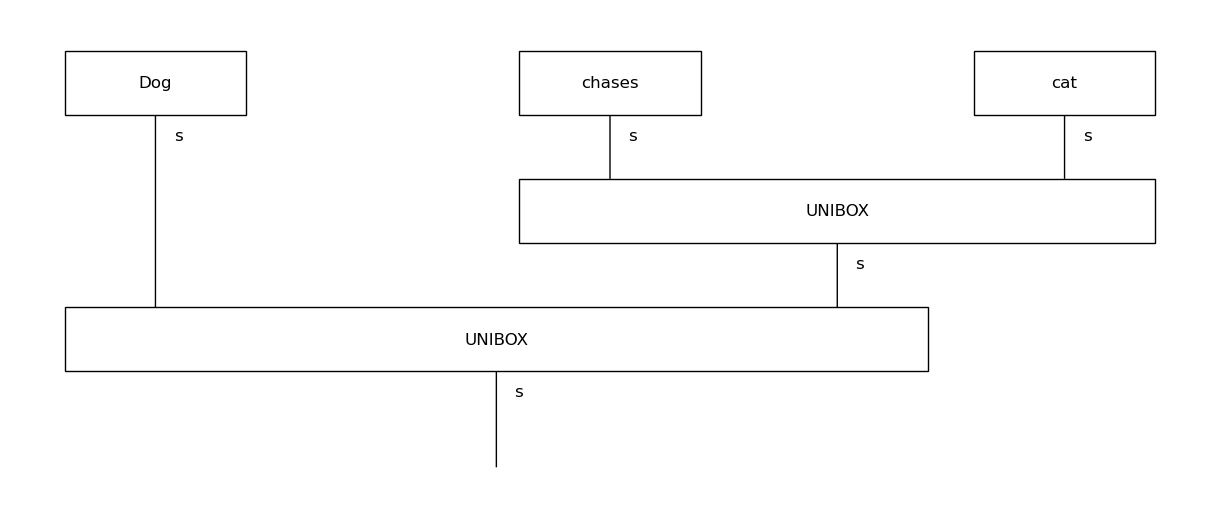} 
    \caption{Tree model string diagram representation for the sentence ``Dog chases cat.''} 
    \label{fig:tree-model} 
\end{figure}


\subsection{Encoding Image Data}
While predefined types exist for sentence composition, incorporating image data requires introducing a new atomic type. To facilitate this, a new box labeled \texttt{"IMAGE"} was added to act as a morphism within the diagram. The domain of this morphism denotes an empty input type and the co-domain is a newly introduced atomic type, \texttt{"image\_type"}, which represents the \textit{n}-dimensional feature vectors derived from images.  This setup enables the diagram to handle and integrate image data by mapping it from a generic input type to a well-defined structure for image data, thus seamlessly incorporating image features into the overall diagrammatic framework.

\begin{figure}[h!] 
    \centering 
    \includegraphics[width=1.0\textwidth]{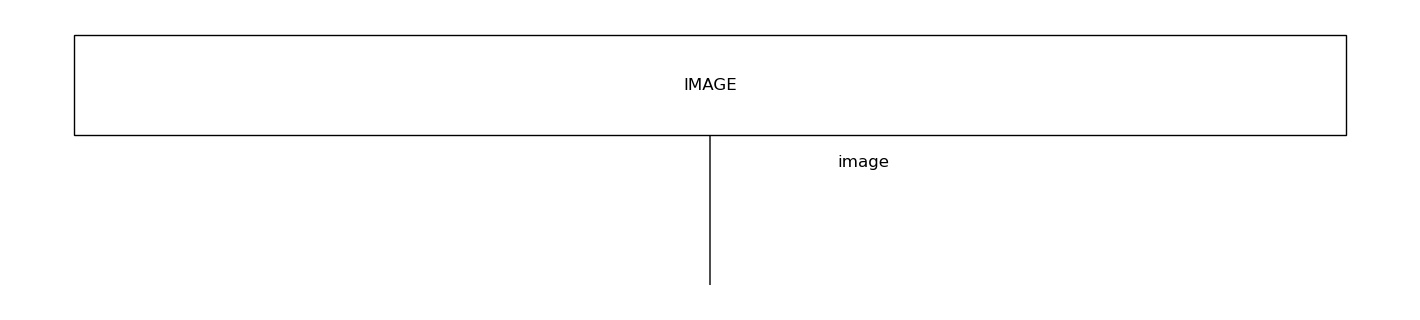} 
    \caption{This diagram illustrates the representation of the atomic type image box for image data.} 
    \label{fig:image-box} 
\end{figure}


\subsection{Unified Diagram Construction}
The concatenation of sentence and image diagrams is achieved through another atomic type, a comparison box. This approach integrates the sentence diagram with the image diagram into a unified diagram. The \texttt{comparison\_box} then processes this combined diagram to produce predicted labels, outputting two probability values. These values indicate the likelihood that the outputs correspond to either a positive or negative classification reflecting the task-specific label’s syntax.

\begin{figure}[h!] 
    \centering 
    \includegraphics[width=1.0\textwidth]{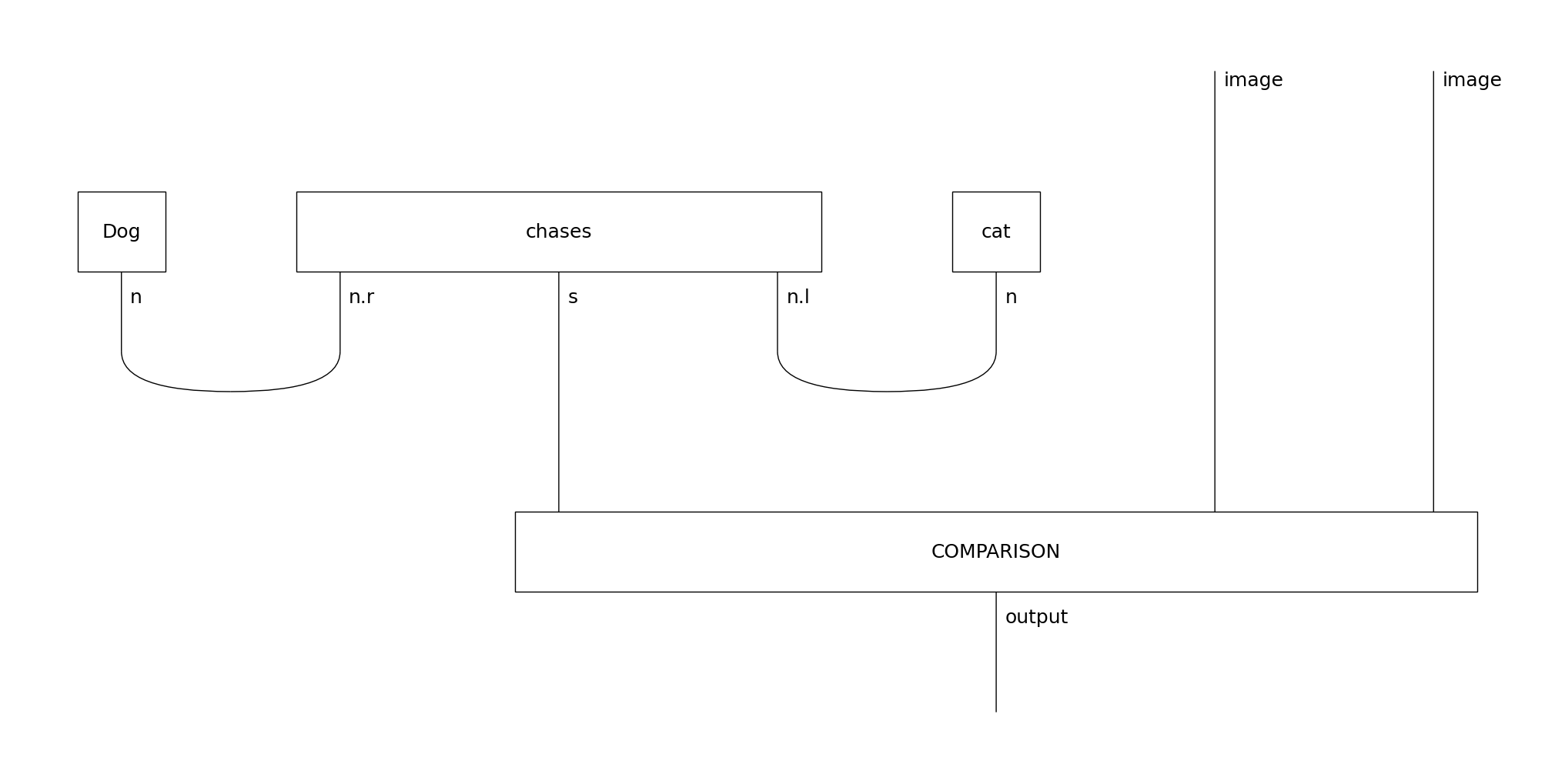} 
    \caption{The diagram shows a unified representation of one sentence and two images merged by a comparison box. The sentence shown uses a syntax-based model; however, this approach can be applied to all the models mentioned above.} 
    \label{fig:conc-diagram} 
\end{figure}


\section{Quantum Circuit Representation}
To convert the merged diagram into a quantum circuit, the process begins by applying an ansatz to the final diagram containing both image and text representations. This ansatz translates the abstract structure of the diagram into a concrete quantum circuit by defining both the number of qubits and the number of layers. It is applied to all components of the diagram except for the image data, which poses a unique challenge due to its high-dimensional nature. To address this, a separate ansatz was specifically developed for the image data. This approach encoded the \textit{n}-dimensional feature vectors, derived from images, into a quantum-compatible format using the \texttt{lambdify} function within \texttt{Lambeq}.

Once the image data has been encoded, it is incorporated into the quantum circuit alongside the components generated from the initial ansatz. This integration is achieved by merging the separately encoded image data with the previously constructed quantum circuit. The image data is integrated at a circuit level, ensuring that it interacts seamlessly with the textual data and other components of the diagram. The result is a unified quantum circuit that accurately reflects the combined information from both textual and image data.

\section{Data Design and Experimental Setup}
To evaluate the effectiveness of the proposed methodology, two distinct datasets were used both of which follow a subject-verb-object (SVO) syntax structure: unstructured dataset and structured dataset.

\subsection{Unstructured Dataset}
\begin{figure}
    \centering
    \includegraphics[width=1\linewidth]{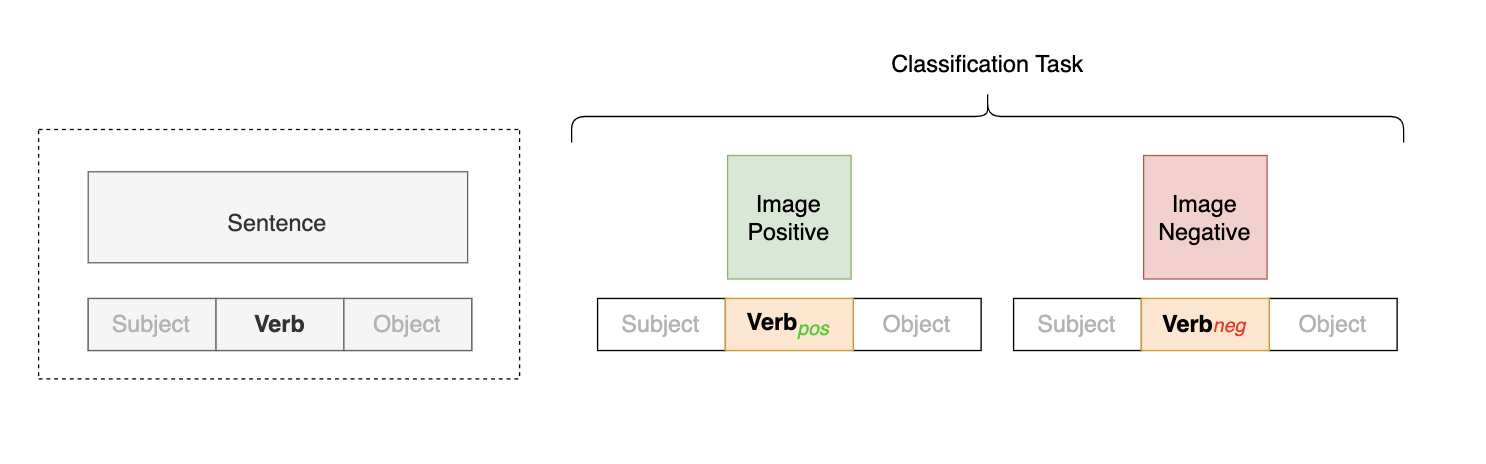}
    \caption{This diagram shows an abstract overview of the classification task for the unstructured dataset.}
    \label{drawio-1}
\end{figure}

\begin{figure}[h!] 
    \centering 
    \includegraphics[width=1.0\textwidth]{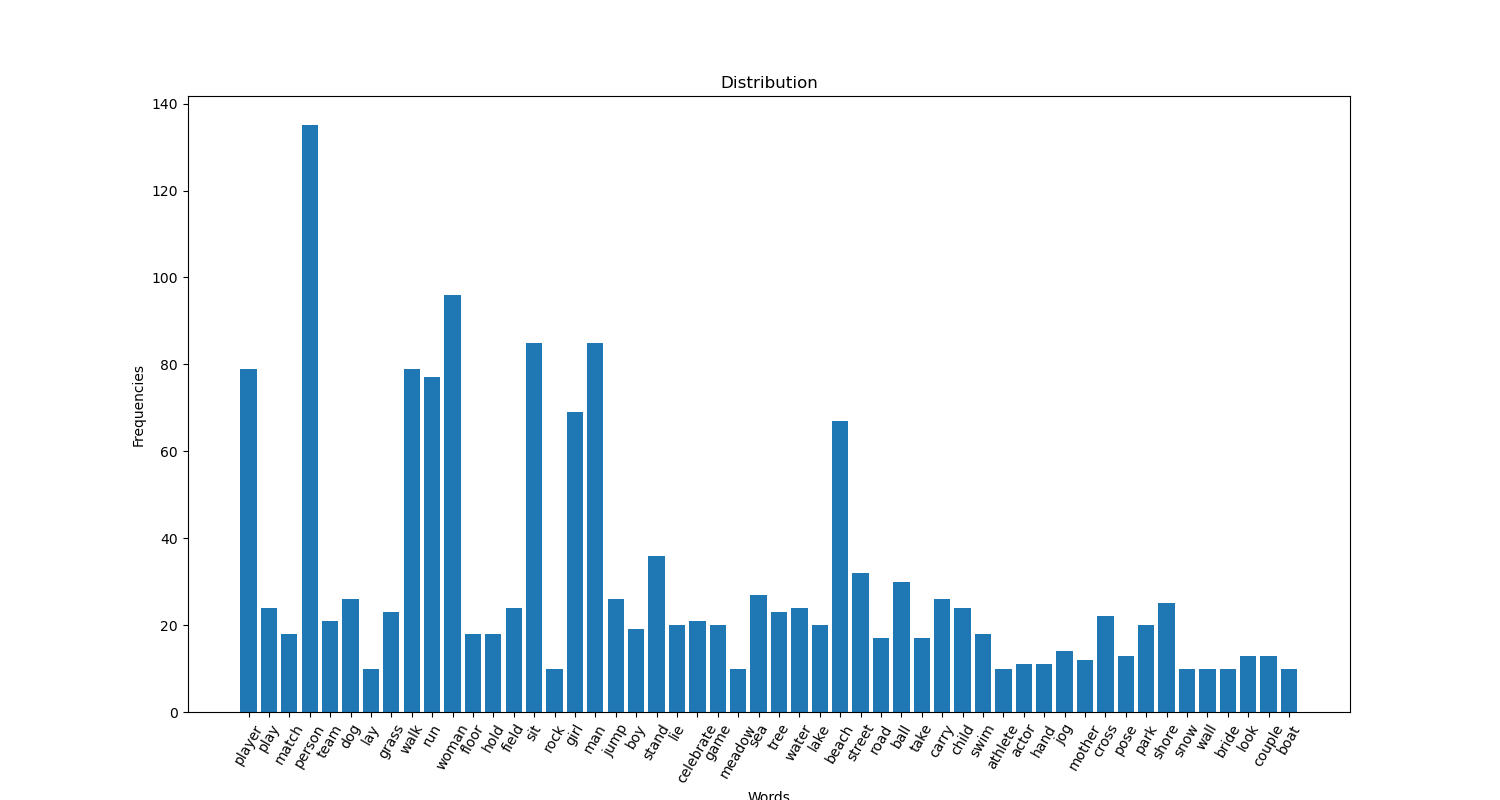} 
    \caption{This graph shows the subject, verb, object distribution for the unstructured dataset.} 
    \label{fig:distribution_3} 
\end{figure}

The unstructured dataset utilizes a large-scale dataset from \cite{Google-SVO}, which introduces a multimodal dataset contains approximately 35,000 entries, where each sentence is paired with two images—one positive and one negative (see Figure \ref{drawio-1}). Each image pair differs by only a single word, ensuring both transparency and consistency when assessing the model’s performance. Due to the processing limitations of the quantum simulator, a subset of approximately 350 entries is used. This subset was selected using automated scripts that ensured all words were repeated and that the distribution was largely balanced, with only a few minor outliers, as shown in Figure \ref{fig:distribution_3}. The data filtering steps were completed using the scripts \texttt{sub-dataset.py} and \texttt{clean\_dataset.py}. Firstly, \texttt{sub-dataset.py} is used to extract a dataset size of approximately 400 entries with an even distribution such that the words found in the dataset are repetitive. This approach ensures that, when the data is divided into training, validation, and test sets, the test set contains new sentences made up of words that the model has seen before. Upon further examination of the dataset, some image URLS were deactivated and no longer in use. Consequently, \texttt{clean\_dataset.py} was used to remove any invalid URLs resulting in a dataset of expected size approximately 350 entries. It is important to note that the dataset maintained its balanced distribution.

\subsection{Structured Dataset}
\begin{figure}
    \centering
    \includegraphics[width=1\linewidth]{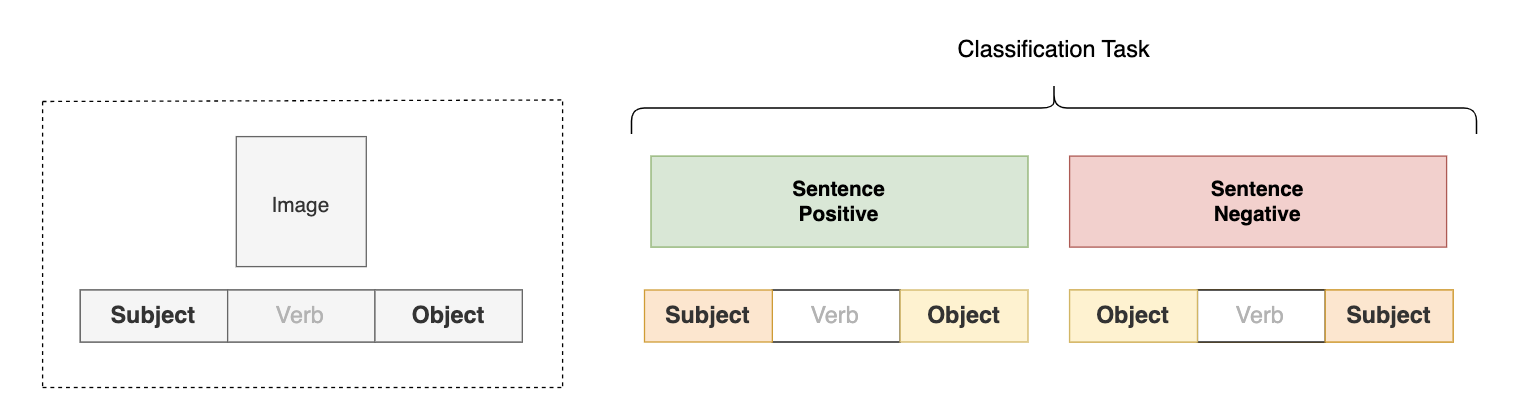}
    \caption{This diagram shows an abstract overview of the classification task for the structured dataset.}
    \label{drawio-2}
\end{figure}

\begin{figure}
    \centering
    \includegraphics[width=0.5\linewidth]{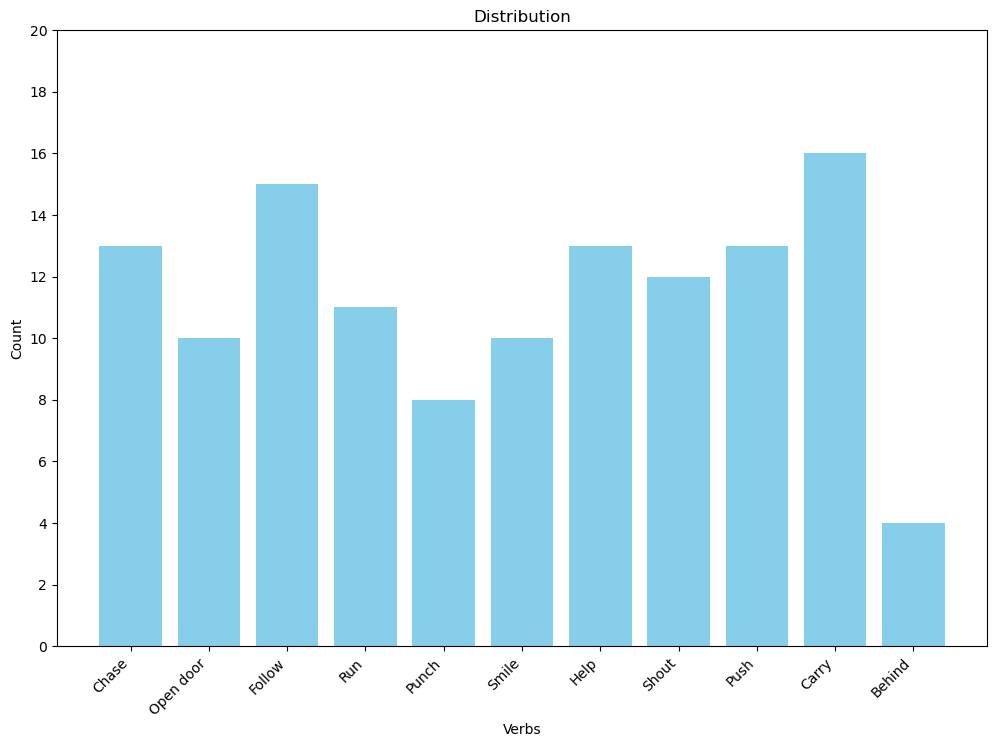}
    \caption{This graph shows the verb distribution for the structured dataset.}
    \label{fig:distribution_dataset2}
\end{figure}

The structured dataset consists of one image paired with two sentence - one positive and one negative. Although a large number of datasets are available for multi-modal applications \cite{DatasetMultimodal}, they neither offer the transparency provided by the subject-verb-object structure nor include interchangeable subjects and objects. To address this, we manually created a small dataset following a simple subject-verb-object syntax structure. For guidance on the dataset size, we referred to original work conducted in \cite{QNLPinPractice}, which used a dataset size of 130. In this dataset, the verb remains constant while the subject and object are interchanged in each sentence pair (see Figure \ref{drawio-2}). The distribution of verbs used is shown in Figure \ref{fig:distribution_dataset2}, ensuring the model's ability to train on various syntactic structures acting on the same verbs. The images were downloaded from Google Search and stored in an image folder within the repository. The dataset was carefully curated to ensure an even distribution of entries and repetitive words.

\newpage
\subsection{Image Feature Extraction}
Both datasets employed the same feature extraction techniques, leveraging Residual Networks (ResNet), a type of convolutional neural network, selected for their ability to capture intricate details and complex patterns. Specifically, the \texttt{Custom16} class in \texttt{feature\_extraction.py} customizes the ResNet-50 model. This class, inheriting from PyTorch’s \texttt{nn.Module}, initialises with a pre-trained ResNet-50 model using weights from ImageNet. The model's architecture is modified to retain all layers except the last two: the global average pooling layer and the fully connected layer. An adaptive average pooling layer is then added to downscale the feature maps to a fixed size, followed by a new fully connected layer to produce an $n$-dimensional feature vector, suitable for quantum circuit encoding.

\chapter{Results}

In this chapter, we present the outcomes of our experiments. We first describe the implementation approach, followed by a detailed presentation of the results obtained from testing these models. 

\section{Implementation}
According to the paper "\textit{QNLP in Practice: Running Compositional Models of Meaning on a Quantum Computer}", which was the first to apply the \texttt{Lambeq} toolkit, four primary approaches to training a model were identified \cite{QNLPinPractice}. These approaches include: a classical tensor network experiment, a quantum simulation where sentences are encoded as quantum circuits, a noiseless quantum simulation using \texttt{qiskit}’s Aer simulator, and a noisy quantum simulation, also executed on \texttt{qiskit}’s Aer simulator. 

For this study, we chose to implement the quantum simulation encoding both text and image data as quantum circuits, with JAX serving as the back-end to manage the training process. This approach allowed us to simulate quantum computations using classical resources, providing a practical environment for developing and testing the model, which can be deployed for further research on actual quantum hardware (see Section \ref{sec:future_work}).

The ansatz for this approach employs the \texttt{Sim14Ansatz}, which pre-defines the number of qubits and layers within the quantum circuits.  After conducting several experiments, the optimal distribution ratio was 1 qubit each for \texttt{AtomicType.NOUN}, \texttt{AtomicType.SENTENCE}, and, when applicable, \texttt{AtomicType.PREPOSITIONAL\_PHRASE}. Additionally, 5 qubits were allocated to \texttt{image\_type}. The number of layers (\texttt{n\_layers}) was set to 1 across all configurations. This distribution resulted in 20 parameters to represent each image, producing a 20-dimensional feature vector during feature extraction. The image vectors were added to the quantum circuit as non-trainable parameters. By keeping the image feature vectors fixed, the image representations remain consistent and unaltered throughout the analysis. This allows us to isolate and study how the static image features influence the results of the sentence parameters processed by the quantum circuit. Additionally, allowing the sentence parameters to be trainable provides the flexibility needed to optimize and adjust the model’s performance. This helps the model learn how to best integrate and interpret the image features in relation to the sentence structure. 

The model in this simulation was built using the \texttt{NumPy} framework and paired with the \texttt{QuantumTrainer} Simulator. We instantiated the model from quantum circuits using the \texttt{NumpyModel.from\_diagrams} function, which was optimized with Just-In-Time (JIT) compilation for improved performance. The choice of JIT compilation allowed for faster execution of operations during training, which is particularly beneficial when dealing with computationally expensive simulations. 

Training was conducted using the Simultaneous Perturbation Stochastic Approximation (SPSA) optimizer, selected for its efficiency in handling high-dimensional optimization problems where gradient calculations can be computationally expensive. However, some drawbacks associated with SPSA are discussed further in Section \ref{sec:limitations}. The Binary Cross-Entropy (BCE) loss function was employed, appropriate for the binary classification tasks for both datasets. Hyperparameters for the optimizer were carefully tuned, with a learning rate (\textit{a}) set at 0.02, a perturbation factor (\textit{c}) of 0.06, and an adjustment parameter (\textit{A}) set to 0.001 times the number of epochs. The adjustment parameter (\textit{A}) controls the step size of perturbations, ensuring that it decreases gradually as training progresses. These settings were chosen to balance the trade-off between convergence speed and training stability. It is important to note that although the same hyperparameters were used for both datasets, the number of epochs and batch sizes were adjusted based on the specific size and characteristics of each dataset. The unstructured dataset, which comprised 350 entries, was trained over 200 epochs with a batch size of 20. In contrast, the structured dataset, consisting of 130 entries based on the dataset size used in \cite{QNLPinPractice}, was trained for 120 epochs with a batch size of 7.

Training labels were extracted directly from the datasets, with labels for the training, validation, and test sets accordingly. The training process was designed to evaluate the model's performance at each epoch, with accuracy as the primary metric. 

\section{Results Table}

This section presents the accuracy results for various methods evaluated on both unstructured and structured datasets. The table below summarizes the average and best accuracy scores achieved by different models, providing a comparative view of their performance.

\definecolor{lightgray2}{rgb}{0.9, 0.9, 0.9}

\begin{table}[h!]
\centering
\renewcommand{\arraystretch}{1.3}  
\setlength{\tabcolsep}{12pt}       
\begin{tabular}{|c|cc|cc|}
\hline
       & \multicolumn{2}{c|}{\textbf{\textit{Unstructured Dataset}}} & \multicolumn{2}{c|}{\textbf{\textit{Structured Dataset}}} \\ \hline \rowcolor{lightgray2}
       & \textbf{Average} & \textbf{Best} & \textbf{Average} & \textbf{Best} \\ \hline
\textbf{Bag of Words}    & 61.88   & 73.58  & 50.00   & 50.00   \\ \hline
\textbf{Cups}   & 56.60   & 58.49   & 49.00   & 60.00   \\ \hline
\textbf{DisCoCat} & \textbf{63.18} & 70.45   & 55.00   & 68.75   \\ \hline
\textbf{Stairs} & 51.70    & 79.25   & 54.00   & 65.00   \\ \hline
\textbf{Tree}   & 60.76   & 64.15   & \textbf{56.00}   & 60.00   \\ \hline
\end{tabular}
\caption{This table displays the accuracy results for both unstructured and structured datasets}
\label{results_table}
\end{table}

\begin{figure}
    \centering
    \includegraphics[width=1.0\linewidth]{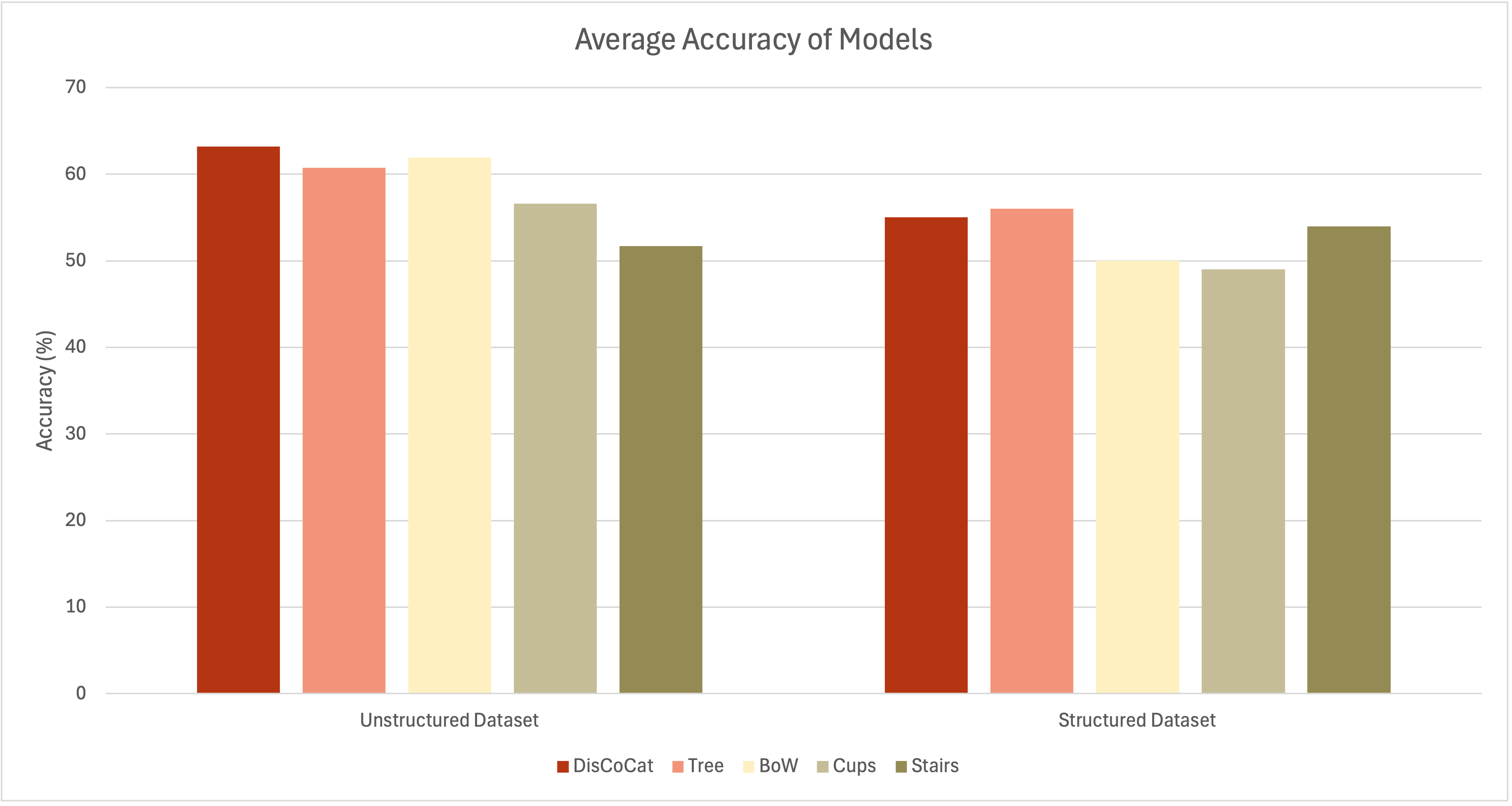}
    \caption{This chart shows the average performance of each model for both datasets.}
\end{figure}

\chapter{Discussion}
This chapter will provide a detailed discussion of the results presented in Table \ref{results_table} and analyze the performance of each model. To ensure a comprehensive evaluation, each model was run five times, and both the average and best results were calculated. Additionally, we will discuss the limitations of this project and suggest ways to improve the methodology and explore further research opportunities. 

\section{Analysis}
The models that exhibited the highest average performance were DisCoCat for unstructured datasets and TreeReader for structured datasets. Both models employ a CCG-based parser, which equips them with a deep understanding of grammatical structure. This approach allows the models to accurately analyse and interpret the syntactic roles of words within a sentence, making them highly effective at capturing the intricacies of language.

The unstructured dataset was designed to evaluate how well models can recognize differences in verb understanding. DisCoCat outperformed the other models, achieving an average accuracy of 63.18\%, which highlights its effectiveness in capturing and distinguishing linguistic elements that other models struggled with. For the structured dataset, DisCoCat achieved the highest individual score of 68.75\% while maintaining the second-best average performance. These results can be attributed to its grammar-aware syntax, which enables it to assign types to various linguistic categories such as nouns and verbs, thereby capturing semantic relationships more effectively. Its strong performance across both datasets demonstrates its robustness in handling various dataset configurations, highlighting its flexibility in classifying different data structures, whether for text or image data.

The TreeReader model also assigns types related to syntactic categories, which are closely tied to the hierarchical structure of the data. This makes the model particularly well-suited for representing relationships within structured datasets. This capability is evident in its performance on the structured dataset, where it achieved the highest average accuracy of 56\%. While this accuracy isn't considered high—an issue further critiqued in Section \ref{sec:limitations}—it serves as a proof of concept that can be further refined, as discussed in Section \ref{sec:future_work}. TreeReader's performance in this context is likely due to its robust handling of word order and syntactic relationships, which enabled it to outperform DisCoCat by better managing variations in subject and object placement.

\begin{figure}
    \centering
    \includegraphics[width=0.8\linewidth]{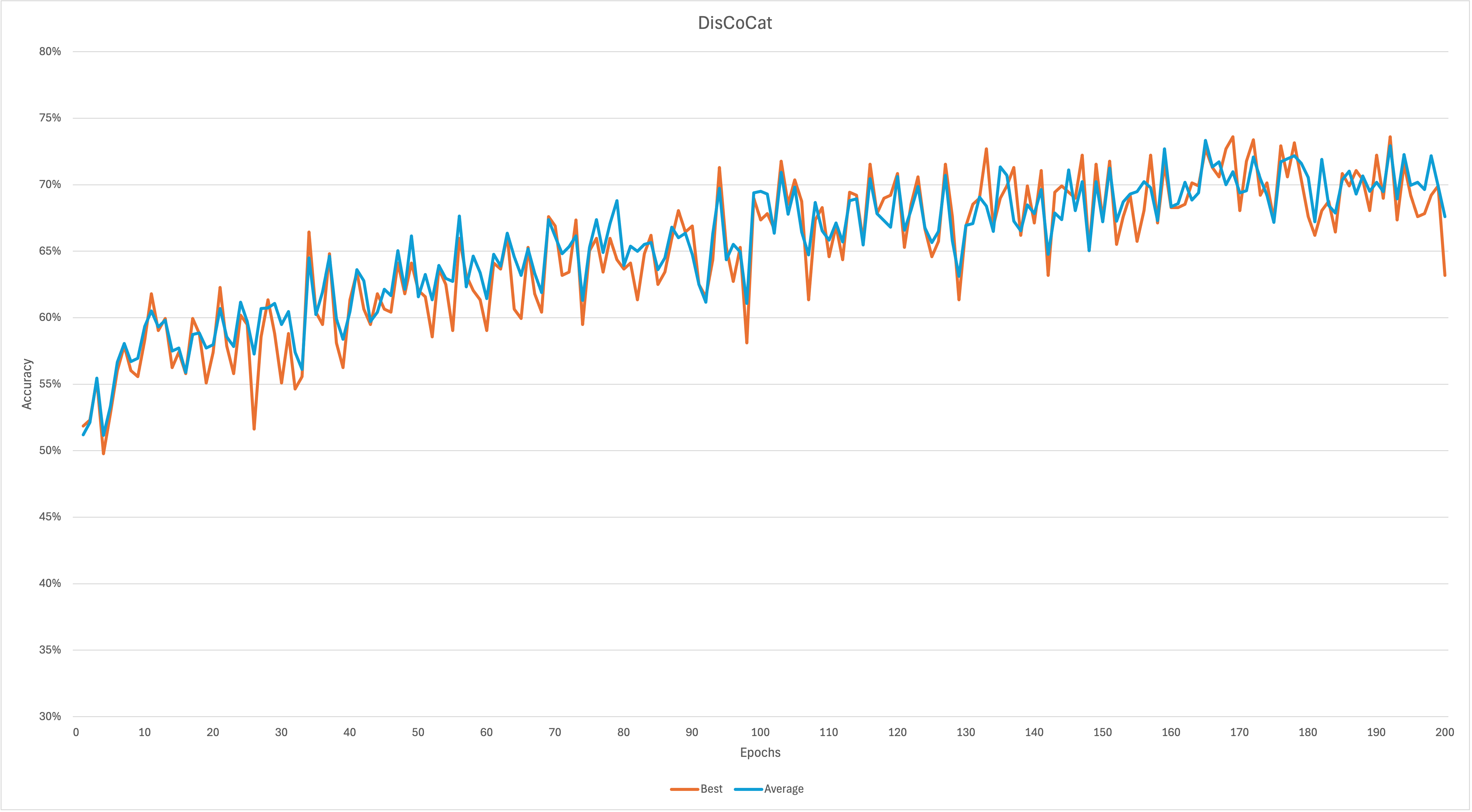}
    \caption{This graph shows the performance of the DisCoCat model during training on 200 epochs for the unstructured dataset.}
\end{figure}

\begin{figure}
    \centering
    \includegraphics[width=0.8\linewidth]{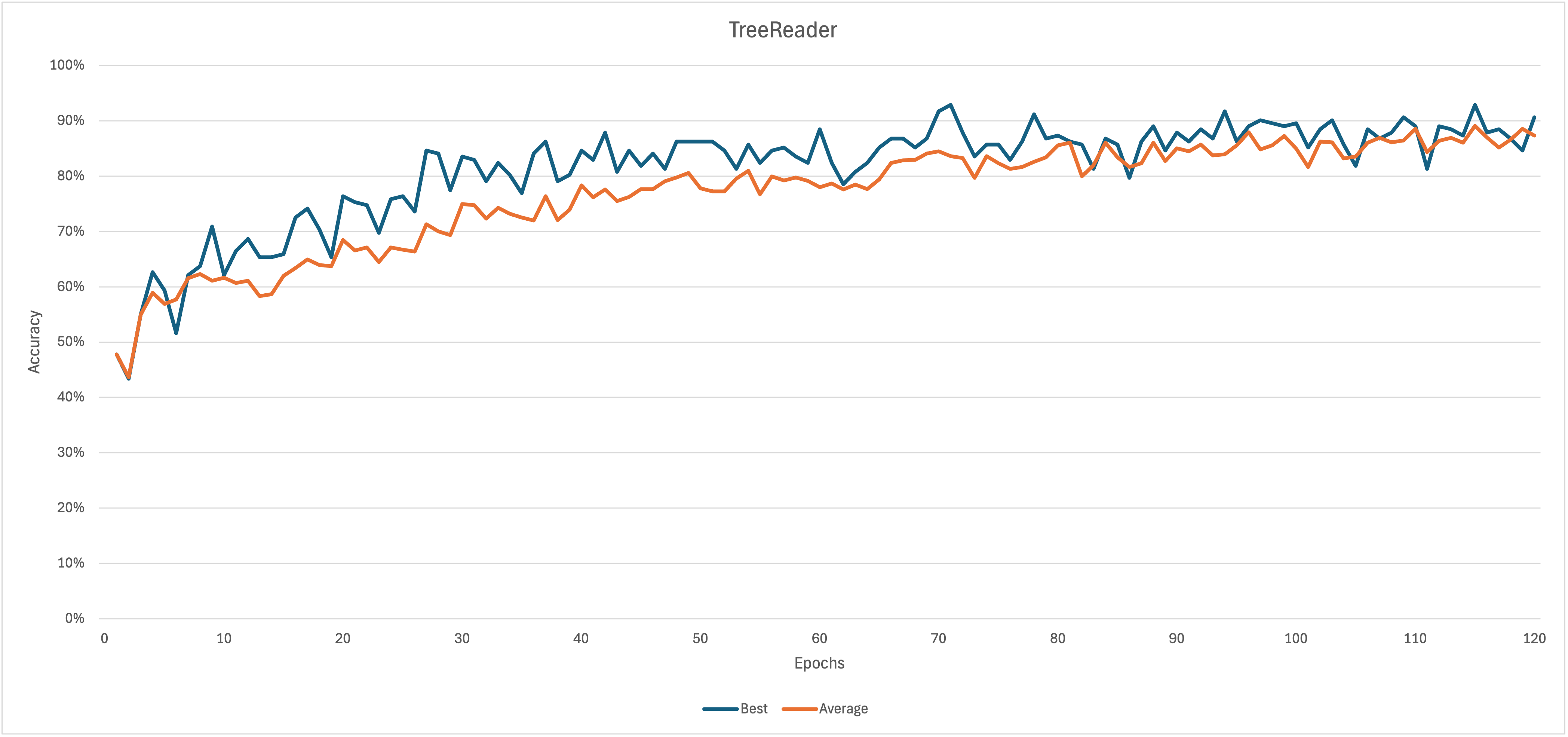}
    \caption{This graph shows the performance of the Tree Reader model during training on 120 epochs for the structured dataset.}
\end{figure}

Word-sequence models like Cups and Stairs process data linearly, without associating word types, leading to distinct performance patterns across both datasets. In the unstructured dataset, the Cups model performed better than Stairs, achieving a higher average accuracy of 56.60\% compared to Stairs’ 51.70\%. This suggests that Cups managed to capture some contextual understanding that Stairs did not. Despite this, both models under-performed relative to others, indicating that their sequential processing approaches struggled with the dataset's variability and complexity. The reliance on sequential order without a strong understanding of grammar or context could have limited their effectiveness in detecting verb usage differences. Cups demonstrated stability during training in the unstructured dataset with only a 2\% difference between its average and best results. This stability contrasts with the structured dataset, where Cups had the lowest average accuracy of 49\% and highest of 60\% indicating a wide range of performance. This significant gap between average and best results suggests that while Cups was relatively stable in the unstructured dataset, it experienced considerable variability in its performance with the structured dataset. This poor performance likely reflects its difficulty managing hierarchical relationships and positional changes between sentence components as designed in the structured dataset. On the other hand, the Stairs model achieved an average accuracy of 54\% in the structured dataset, which was slightly better than Cups but still not optimal. This result indicates that while Stairs' sequential approach had some improvements in dealing with structured data, it struggled with hierarchical relationships and positional changes, similar to Cups.

\begin{figure}
    \centering
    \includegraphics[width=1.0\linewidth]{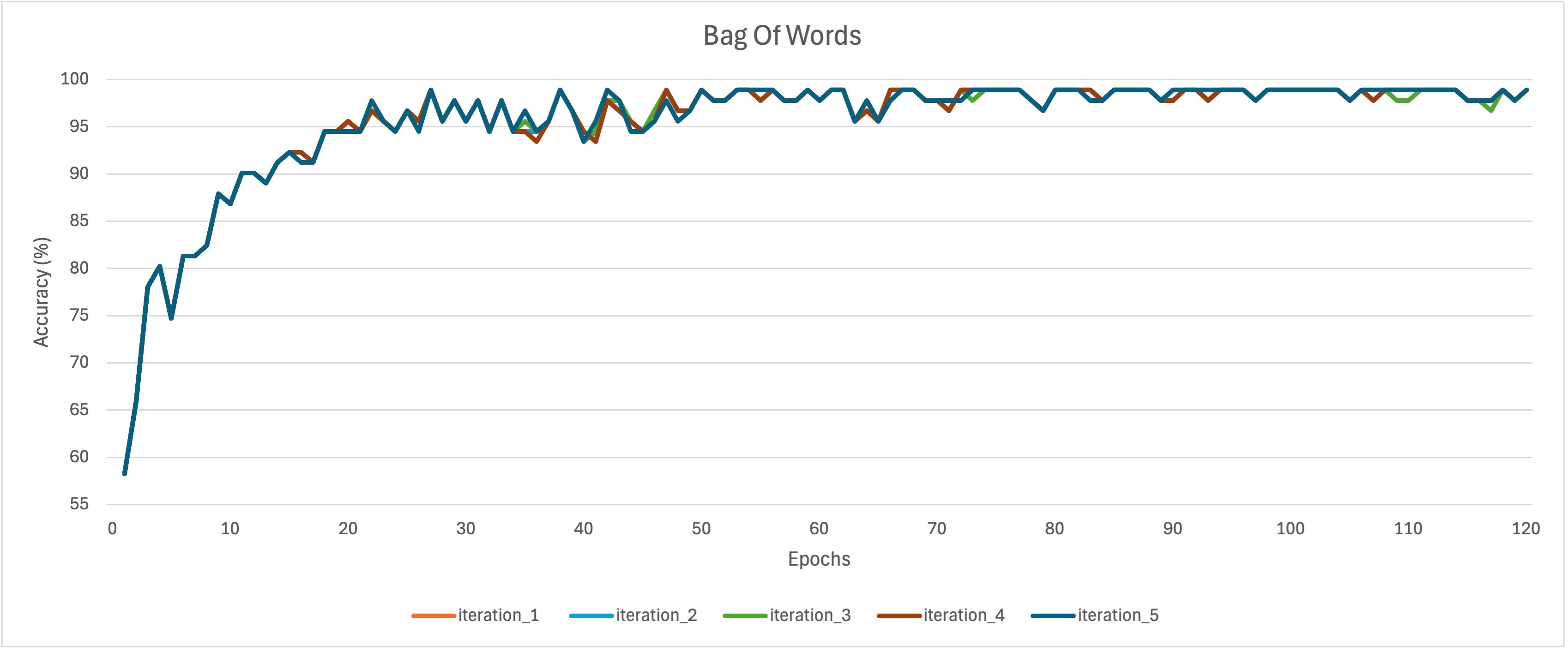}
    \caption{This figure shows the accuracy metric for the Bag-of-Words model during training with the structured dataset.}
    \label{Fig: BoW_ACC}
\end{figure}

\begin{figure}
    \centering
    \includegraphics[width=0.8\linewidth]{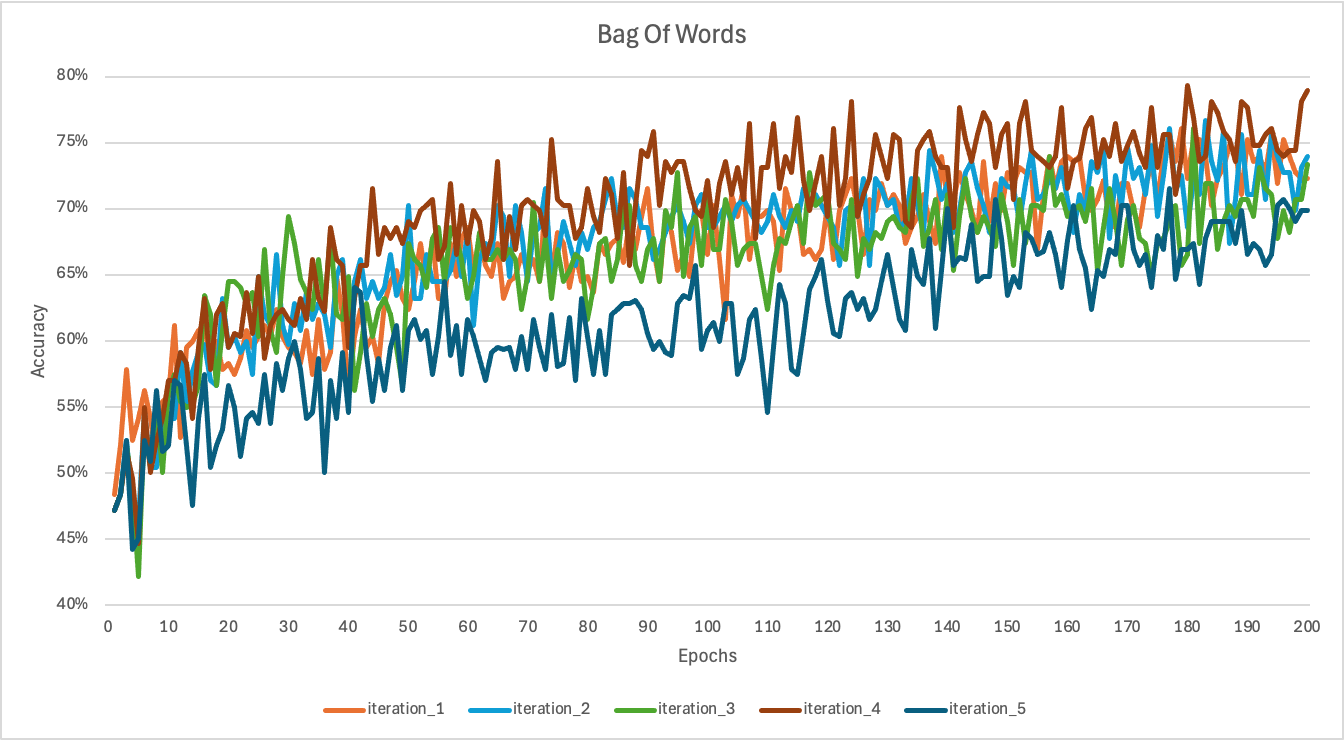}
    \caption{This figure shows the accuracy metric for the Bag-of-Words model during training with the unstructured dataset}
    \label{fig: BoW_ACC_dataset1}
\end{figure}

The Bag-of-Words model provides an interesting perspective on performance across both datasets. For the unstructured dataset, it achieved relatively high performance, securing the second-best scores with only a 2\% variation. This demonstrates its ability to identify and differentiate verbs, highlighting its strength in basic word recognition. In contrast, the model's performance on the structured dataset was notably poor, differing by just 1\% from the lowest average result recorded by Cups. This poor performance is likely due to its lack of syntactic understanding, which is crucial for handling the positional relationships and interchange of subjects and objects as depicted in the structured dataset. This is further illustrated in Figure \ref{Fig: BoW_ACC}, which demonstrates a more consistent performance across all iterations for the structured dataset. In contrast, Figure \ref{fig: BoW_ACC_dataset1} reveals significantly more noise in the performance metrics for the unstructured dataset.

The results of the analysis align with the central hypothesis of this study, which states that syntactic and structure-aware models outperform those that ignore grammar and sentence composition. Both DisCoCat and TreeReader demonstrated the benefits of leveraging grammatical structures, as they achieved the highest performance across unstructured and structured datasets, respectively. DisCoCat's noteable accuracy in distinguishing verb usage in the unstructured dataset supports the sub-hypothesis that verb composition and syntactic differentiation are crucial for model performance. This reinforces the idea that models capable of interpreting grammatical relationships, such as subject-object variations and verb interaction, excel in real-world linguistic tasks. Furthermore, TreeReader's results with structured data highlights the importance of hierarchical understanding, as it outperformed other models by managing word order and syntactic relationships.

In contrast, the Bag-of-Words model, which lacks syntactic awareness, further justifies the central hypothesis. Its relatively uniform performance across training iterations reflects its inability to differentiate between sentence structures in the structured dataset. Without a clear understanding of grammatical relationships or positional changes between sentence components, the Bag-of-Words model treated sentences in a similar manner regardless of their syntactic differences. This outcome emphasizes the importance of structure in achieving higher accuracy, particularly in tasks requiring a deep understanding of sentence composition, as supported by the performance of grammar-aware models like DisCoCat and TreeReader.

\begin{figure}
    \centering
    \includegraphics[width=1.0\linewidth]{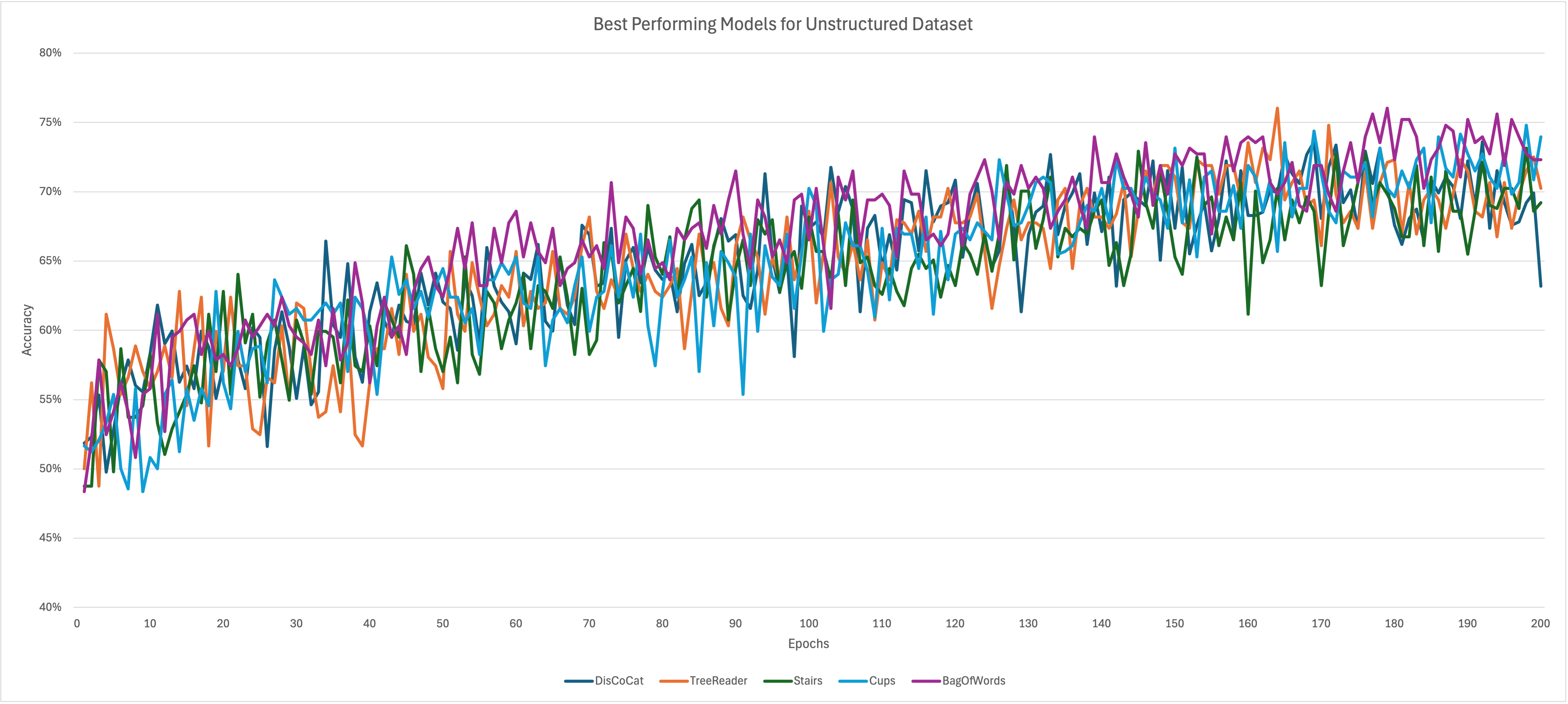}
    \caption{This figure shows the accuracy metrics for best performing models for unstructured dataset.}
    \label{fig:Dataset1_best}
\end{figure}

\begin{figure}
    \centering
    \includegraphics[width=1.0\linewidth]{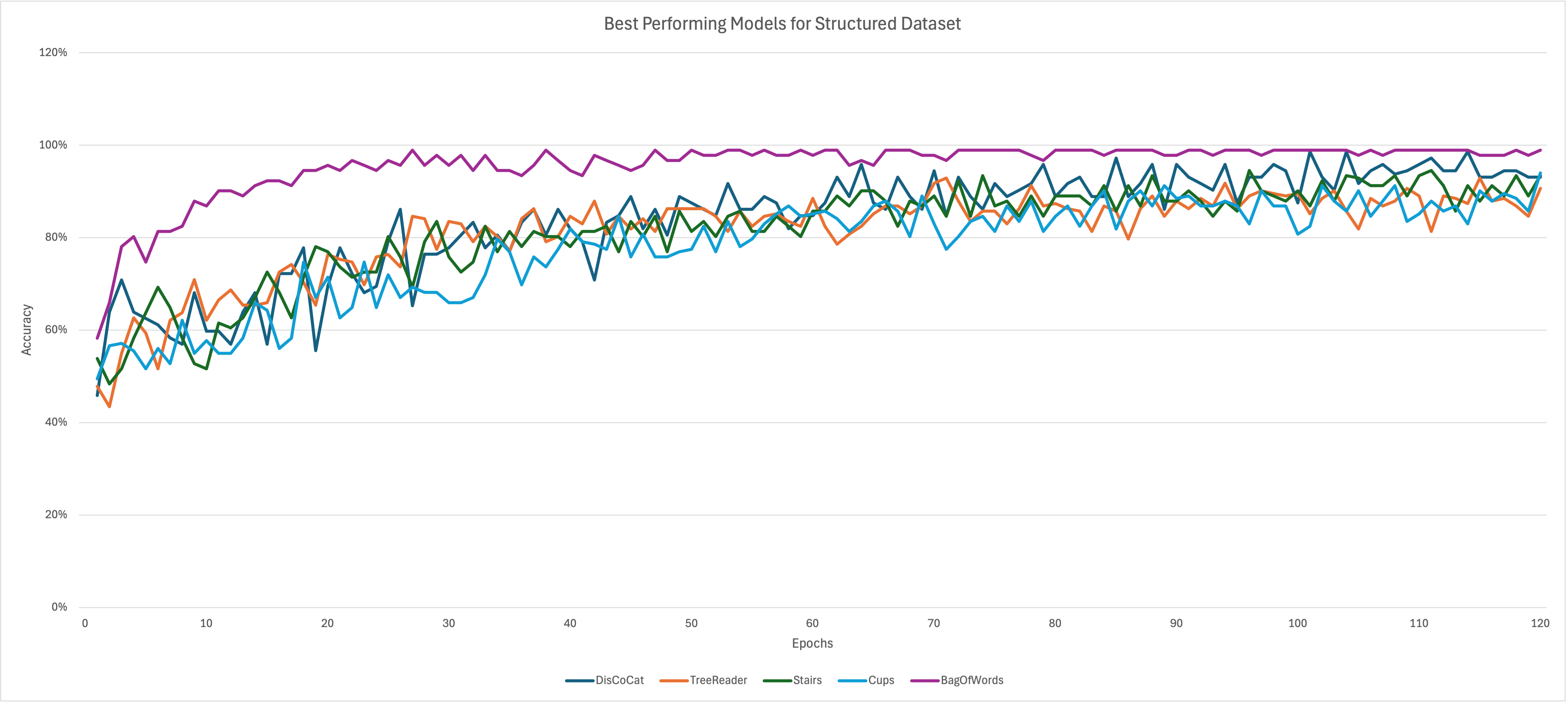}
    \caption{This figure shows the accuracy metrics for best performing models for structured dataset.}
\end{figure}

\newpage
\section{Limitations}
\label{sec:limitations}
Despite the promising results, several limitations impacted the project's outcomes. One key issue was the reliance on a Numpy-based model, which relied on a SPSA optimizer. This optimizer, which approximates the gradient by perturbing all parameters simultaneously, introduced noise and led to the vanishing gradient problem (as discussed in Section \ref{visionlearning}).  To address this, we used a small learning rate of 0.001. Although the model showed some signs of learning, particularly with the unstructured dataset (see Figure \ref{fig:Dataset1_best}), it had not yet reached a stable convergence point, likely affecting overall performance and introducing significant noise during training. Employing a more advanced model could have mitigated these challenges and led to better results. However, due to the computational limitations of the project and available resources, we were unable to utilise more sophisticated models. These constraints required the use of simpler, less resource-intensive approaches, which, while functional, may have compromised the potential for achieving optimal performance.

The project's computational constraints limited the dimensionality of the image vector representations. While ResNet was used to extract crucial features, there was a significant difference in image representation between the 20-dimensional vectors used in this research and the 2048-dimensional vectors employed in classical models. The higher-dimensional vectors capture essential features such as edges and corners more effectively, whereas the lower-dimensional vectors considerably reduced image representation quality, likely hindering or delaying the training process. Despite this limitation, many models averaged more than 60\% accuracy, indicating that the model is learning effectively. Some models even achieved accuracy rates as high as 79\%. Although a higher-dimensional feature vector would likely have improved the results, this finding highlights the interesting potential of quantum architecture in image representation. It suggests that quantum models may capture image features in ways that classical models have not been able to achieve.

Another notable limitation of the study was the size of the dataset. In comparable research involving classical classifiers, larger datasets — often exceeding 15,000 samples as demonstrated in \cite{corefernce} — are commonly employed. Such extensive datasets are crucial for training models effectively, which can significantly improve their performance. A larger dataset would not only provide more training examples but also enhance the model's ability to generalize from these examples. Furthermore, a larger dataset would enable a more extended training period across multiple epochs, allowing it to capture complex patterns within the data and learn more features, resulting in a more robust and reliable model performance. 

\section{Future Work}
\label{sec:future_work}
For future work, multiple approaches can be explored to enhance the research and address current limitations. Firstly, employing larger datasets and extending training durations could significantly improve model performance and generalization. Additionally, incorporating two benchmarks—one using a random dataset for comparative evaluation and another against state-of-the-art models—will provide a more comprehensive assessment of the model's performance. Utilising advanced quantum computing frameworks, such as the Pennylane model, could offer new insights and improvements in image representation. Furthermore, conducting more controlled experiments using the same dataset across different models will help determine if performance improvements are more pronounced in sentence classification or image classification tasks. Additionally, exploring higher-dimensional feature vectors and optimizing feature vector representations by fine-tuning on classification tasks could further enhance the effectiveness of the models. Finally, investigating the deployment of models on quantum hardware represents an intriguing area of research. Although current noise levels resulting from NISQ devices may not directly impact performance improvements, it will be interesting to see how these models perform as quantum technology advances.
 
\chapter{Conclusion}

This thesis introduced a framework for applying MQNLP, marking the first investigation of image and text data on a quantum simulator. Utilizing \texttt{Lambeq}, a high-level Python toolkit, we designed and trained multimodal image-text quantum circuits. The study involved two experiments with distinct datasets tailored for classification tasks: structured and unstructured. We analysed the performance of various compositional models—syntax-based, bag-of-words, word-sequence, and tree-based—on these datasets to assess their effectiveness in understanding and processing linguistic syntax.

The study highlighted that syntax-based and grammar-aware models, particularly DisCoCat and TreeReader, consistently outperformed others. Both models demonstrated that an understanding of syntax and structure significantly enhanced performance. DisCoCat achieved an average accuracy of 63.18\% on unstructured data and a peak of 68.75\% on structured data, while TreeReader performed best on structured data with an average accuracy of 56\%. These remarkable results indicate that the models’ deep grammatical understanding and ability to interpret syntactic roles significantly enhance their effectiveness and overall performance.

In contrast, the Bag-of-Words model underperformed on structured data, as expected, due to its lack of syntactic differentiation. Its performance was consistent across iterations, reflecting its inability to distinguish between sentence structures. Nevertheless, it performed relatively well on unstructured data, achieving a second-best average accuracy of 61\%. This suggests that while it lacks syntactic insight, it could still capture some verb-word usage and their representation in image vectors.

Sequence models, like Cups and Stairs, did not perform as well. These models struggled with both datasets because their linear processing lacked the necessary understanding of word composition and structure. The sequential order of words was insufficient for capturing complex syntactic relationships, which impacted their overall effectiveness.

Our main hypothesis—that structure-aware models would perform better on linguistic data—was supported, with DisCoCat and TreeReader validating the importance of syntactic and grammatical understanding. The sub-hypothesis, focusing on verb composition and understanding, was also affirmed, especially in unstructured data where the ability to differentiate verbs led to higher accuracy. This showed that models equipped with a deep understanding of structure and verb relationships could perform better in tasks requiring precise language processing. The study extended these hypotheses to real-world data, such as image-text pairs, and found that structure-aware models showed promise beyond traditional language tasks. This broader applicability suggests that understanding syntax and structure can be beneficial in multimodal contexts, paving the way for future research and application in more complex data scenarios.

The results highlight the significant promise of MQNLP, with the potential to surpass classical methods. Our experiments, even with smaller image vectors, achieved accuracy's close to those of classical models. This indicates that, with further development, quantum approaches could outperform traditional ones. The findings emphasize the need to explore multimodal applications in the context of quantum computing, extending research beyond classical methods and leveraging current quantum-enhanced technologies.

\chapter{Summary}

\section{Project Title}
The title of this thesis is "\textit{Multimodal Quantum Natural Language Processing: A Novel Framework for Using Quantum Methods to Analyse Real Data}", supervised by Dr. Mehrnoosh Sadrzadeh.

\section{Project Overview}

Despite the progress in QNLP, its application has been limited to textual data, unlike classical NLP that integrates multiple data modalities. This research advances the field of Multimodal QNLP (MQNLP) by focusing on image-text classification. It introduces a novel framework that incorporates both images and text, leveraging the \texttt{Lambeq} toolkit to design multimodal image-text quantum circuits and train their parameters.  The study evaluates various compositional methods—including syntax-based, bag-of-words, word-sequence, and tree-based models—on the developed datasets. 

\begin{itemize}
    \item \textbf{Experiment 1}: Analyzes verb-based differences using sentences paired with positive and negative images, evaluating how well models distinguish these differences.
    \item \textbf{Experiment 2}: Assesses the impact of sentence structure and word order by comparing sentences with interchanged subjects and objects, paired with a single image.
\end{itemize}

\section{Collaborative Efforts}
Throughout this project, I consulted informally with Dr. Dimitri Kartsaklis at Quantinuum, gaining valuable insights and feedback on various aspects of the research. I also received assistance from Nikhil Khatri, which helped address technical roadblocks and refine certain elements of the project. Additionally, I participated in the reading group led by Prof. Mehrnoosh Sadrzadeh at UCL where we reviewed relevant literature and discussed advancements in the field. These interactions significantly contributed to the development of the framework and enhanced the quality of the research.

\section{Project Results}
The results of the analysis align with the central hypothesis that syntactic and structure-aware models outperform those that ignore grammar and sentence composition. Specifically:
\begin{itemize}
    \item \textbf{DisCoCat} demonstrated notable accuracy in distinguishing verb usage within the unstructured dataset, supporting the idea that verb composition and syntactic differentiation are critical for model performance.
    \item \textbf{TreeReader} excelled with the structured dataset by effectively managing word order and syntactic relationships, underscoring the importance of hierarchical understanding in achieving high performance.
    \item \textbf{Bag-of-Words}, which lacks syntactic awareness, showed relatively uniform performance, highlighting its limitation in differentiating sentence structures. 
\end{itemize}

The promising results of this study emphasises the potential of MQNLP and suggest that further exploration in this area could yield valuable insights and advancements in the field.

\addcontentsline{toc}{chapter}{Bibliography} 
\bibliography{References}

\begin{appendices}

\chapter{Sample Datasets}
\begin{table}[h!]
    \centering
    \begin{tabular}{|l|l|l|c|c|}
        \hline
        \textbf{Sentence} & \textbf{Image1} & \textbf{Image1} & \textbf{Label1} & \textbf{Label2} \\
        \hline
        Players play golf. & \includegraphics[width=0.1\textwidth,height=0.1\textheight,keepaspectratio]{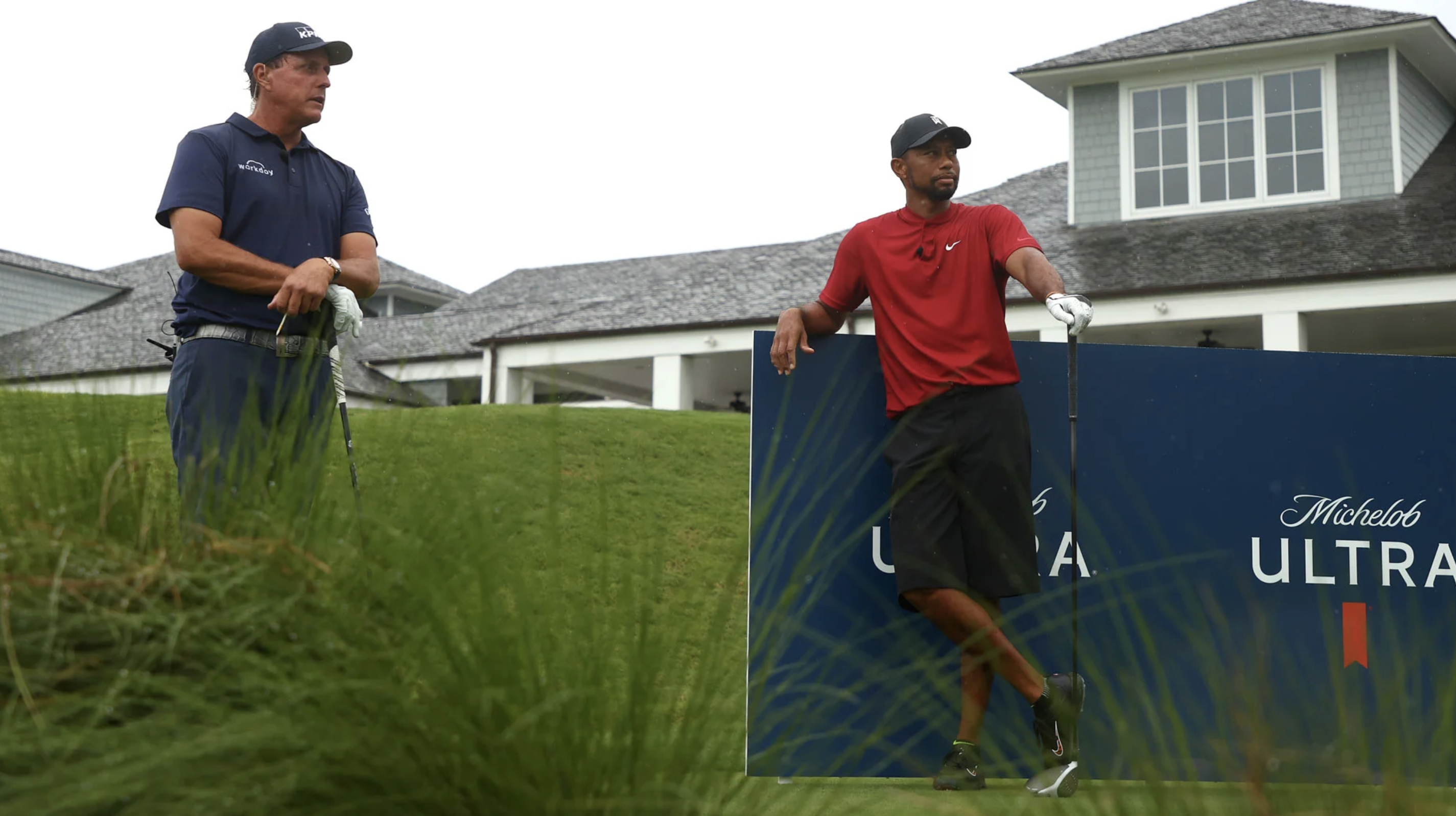} & \includegraphics[width=0.1\textwidth,height=0.2\textheight,keepaspectratio]{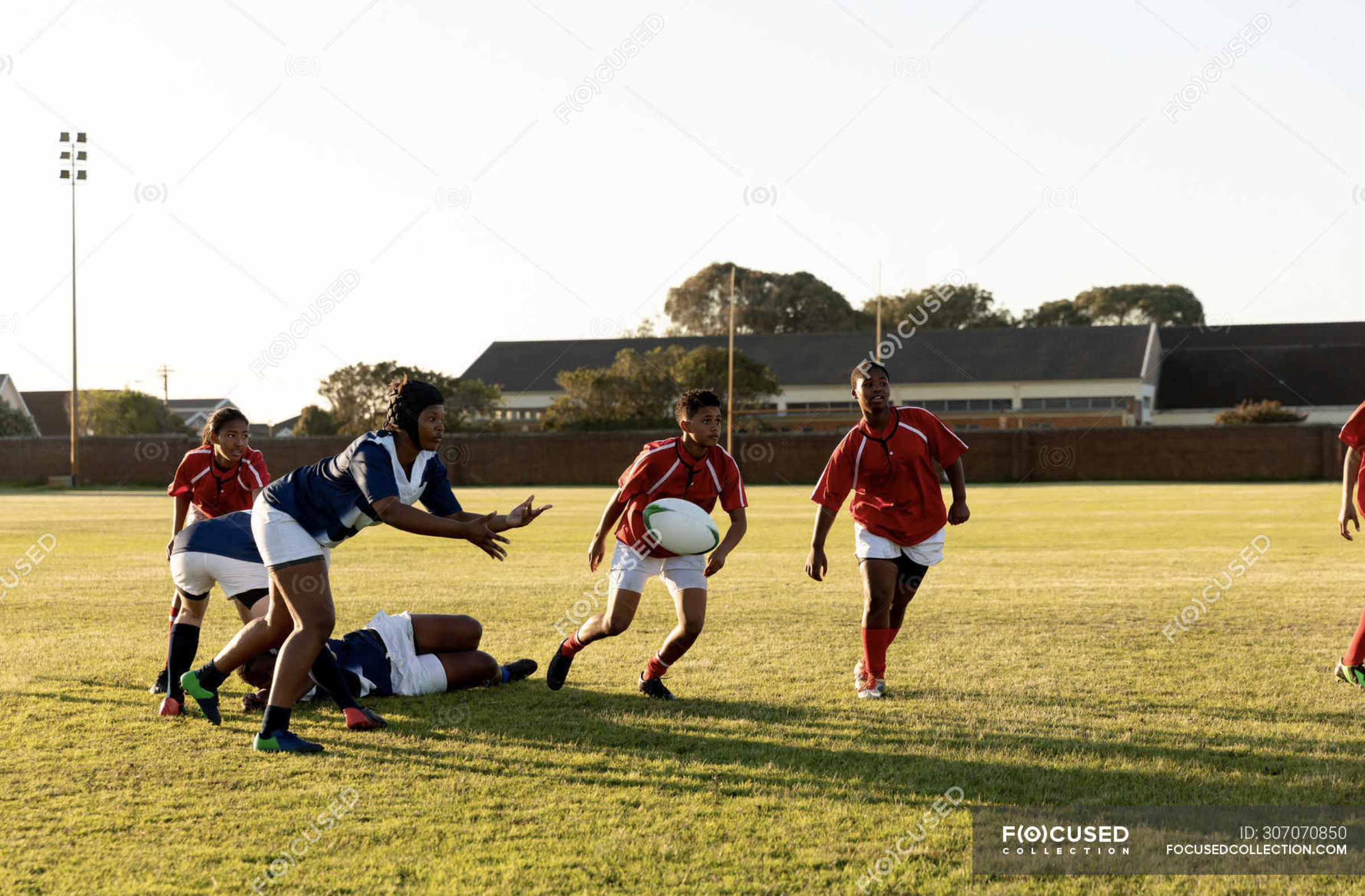} & 0 & 1 \\
        \hline
        Dog lays on grass. & \includegraphics[width=0.1\textwidth,height=0.2\textheight,keepaspectratio]{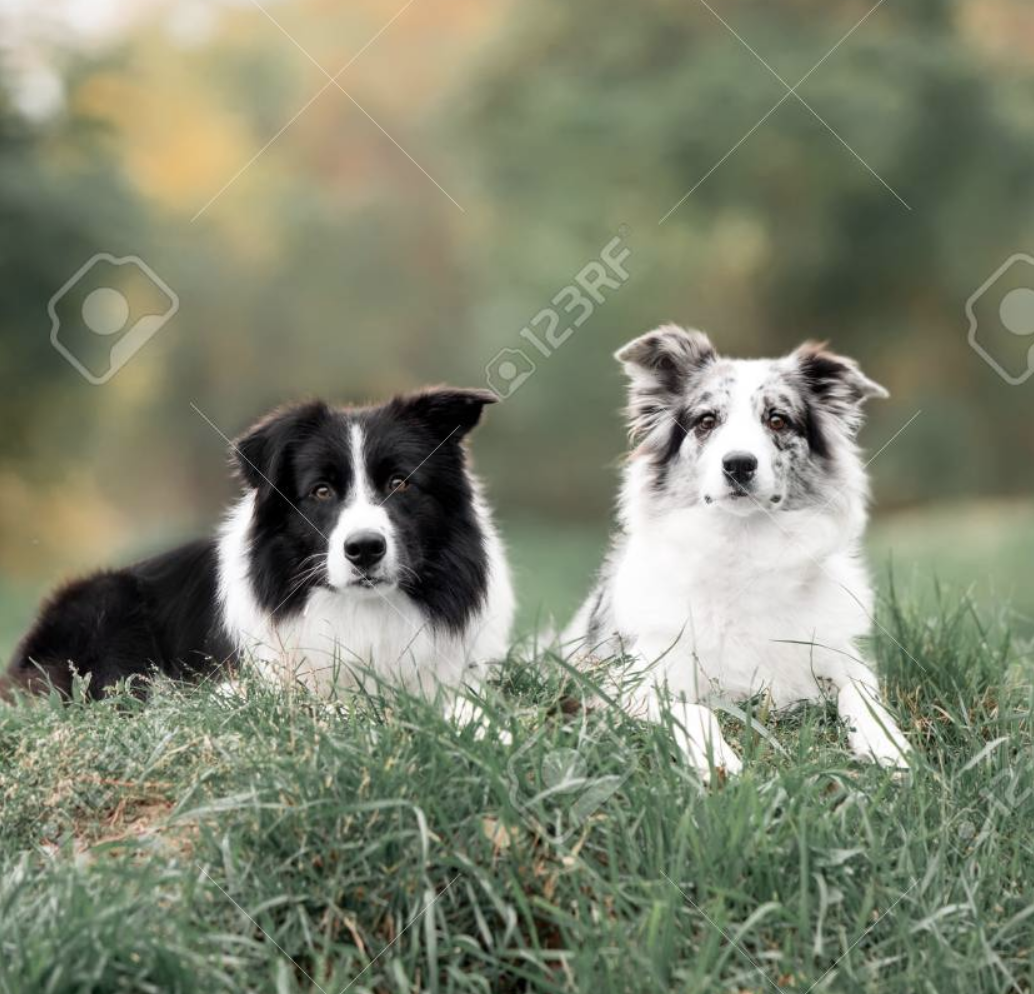} & \includegraphics[width=0.1\textwidth,height=0.2\textheight,keepaspectratio]{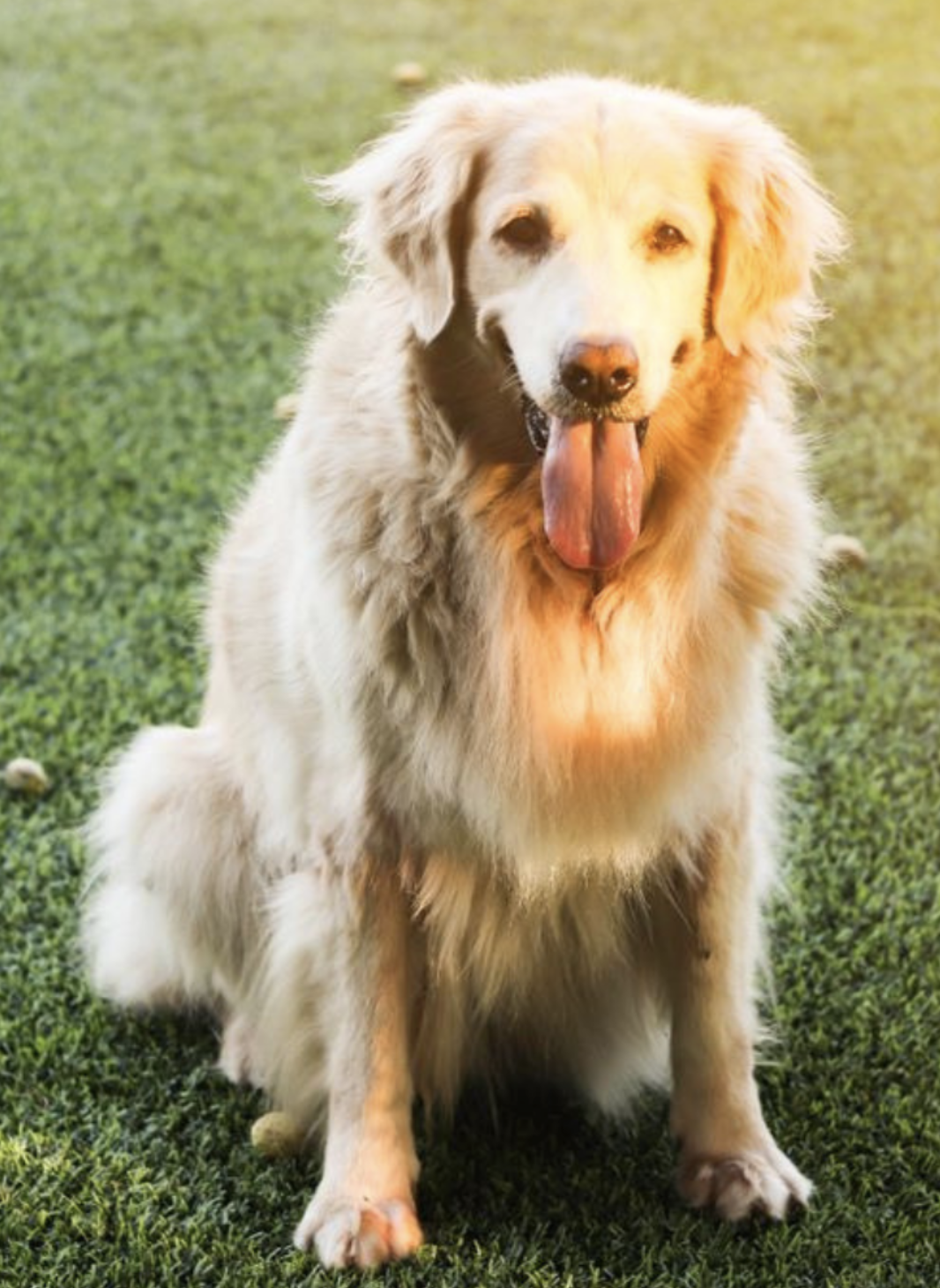} & 0 & 1 \\
        \hline
        Person runs as a player 
        
        in competitive sport. & \includegraphics[width=0.1\textwidth,height=0.2\textheight,keepaspectratio]{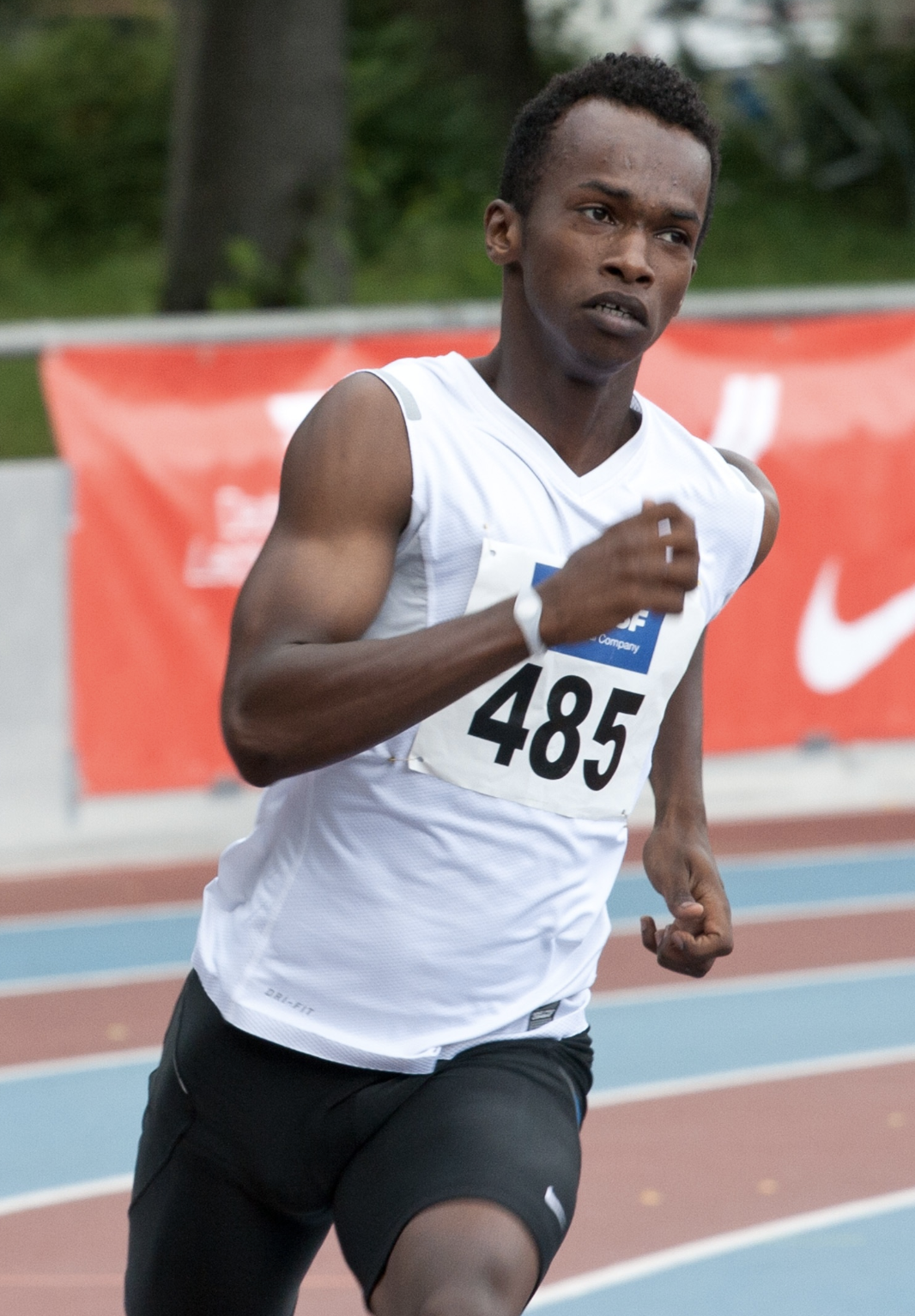} & \includegraphics[width=0.1\textwidth,height=0.2\textheight,keepaspectratio]{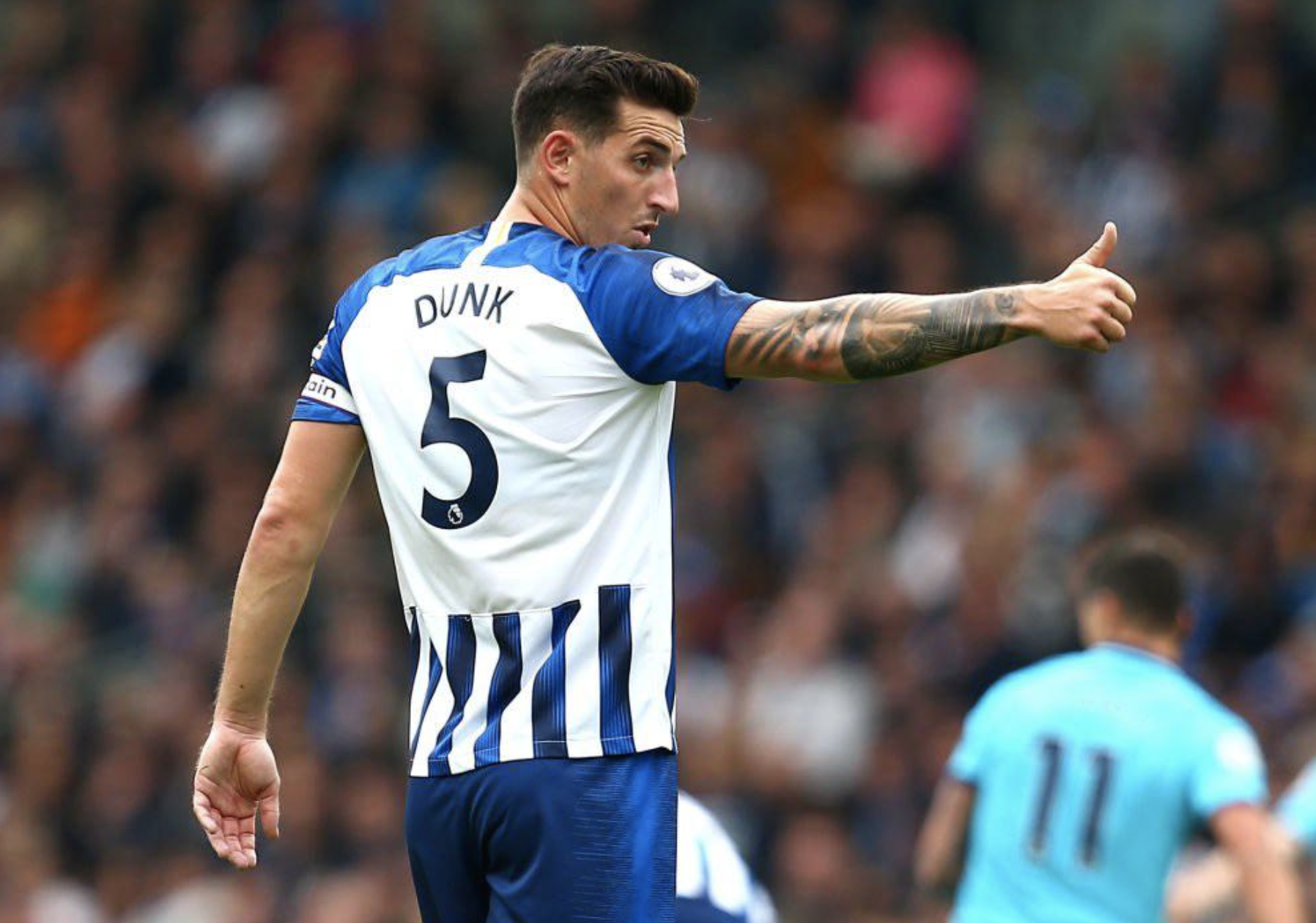} & 0 & 1 \\
        \hline
    \end{tabular}
    \caption{Sample entries from the unstructured dataset.}
    \label{tab:sample_entries}
\end{table}

\begin{table}[h!]
    \centering
    \begin{tabular}{|l|l|l|c|c|}
        \hline
        \textbf{Sentence 1} & \textbf{Sentence 2} & \textbf{Image} & \textbf{Label} \\
        \hline
        Cat chases mouse. & Mouse chases cat. & \includegraphics[width=0.1\textwidth,height=0.07\textheight,keepaspectratio]{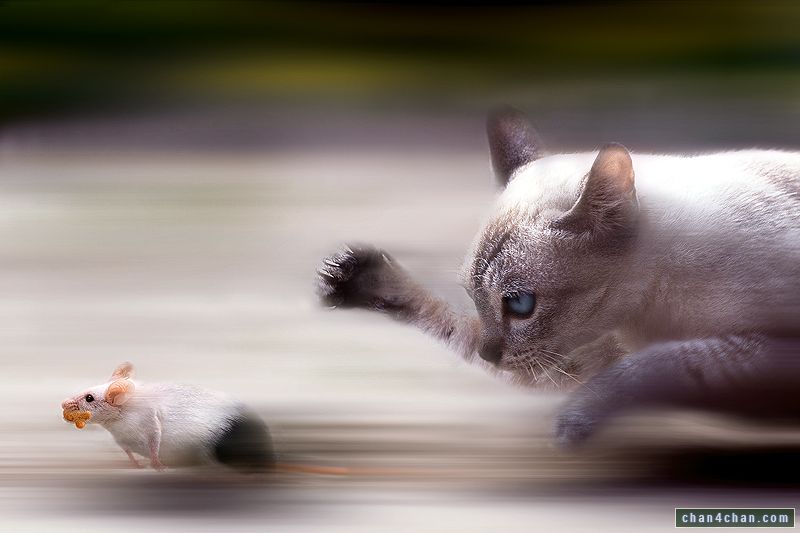} & 0 & 1 \\
        \hline
        Lion chases zebra. & Zebra chases lion. & \includegraphics[width=0.1\textwidth,height=0.07\textheight,keepaspectratio]{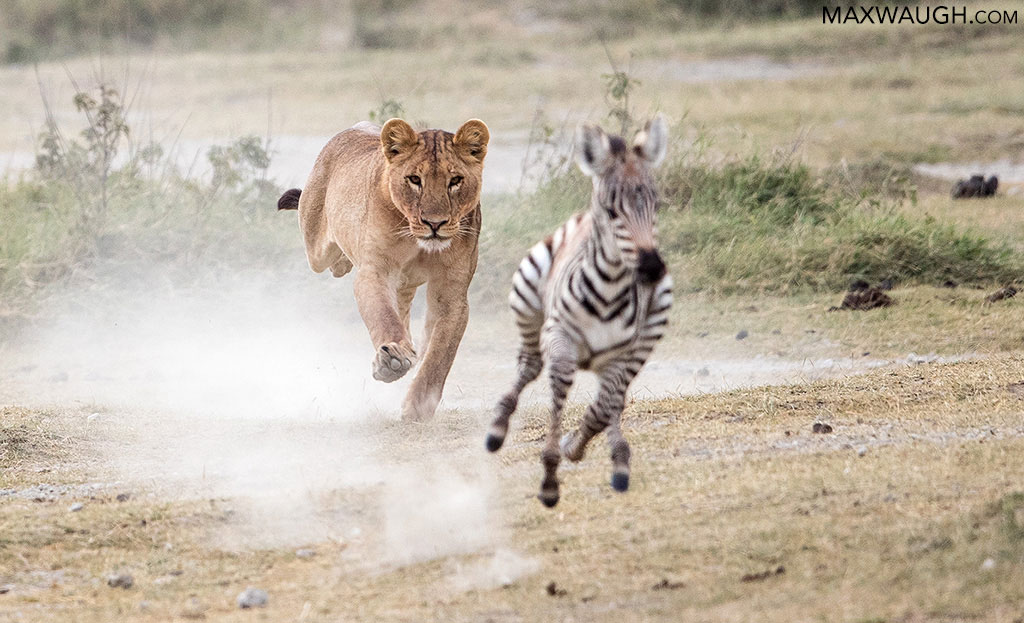} & 0 & 1 \\
        \hline
        Cheetah chases gazelle. & Gazelle chases cheetah. & \includegraphics[width=0.1\textwidth,height=0.07\textheight,keepaspectratio]{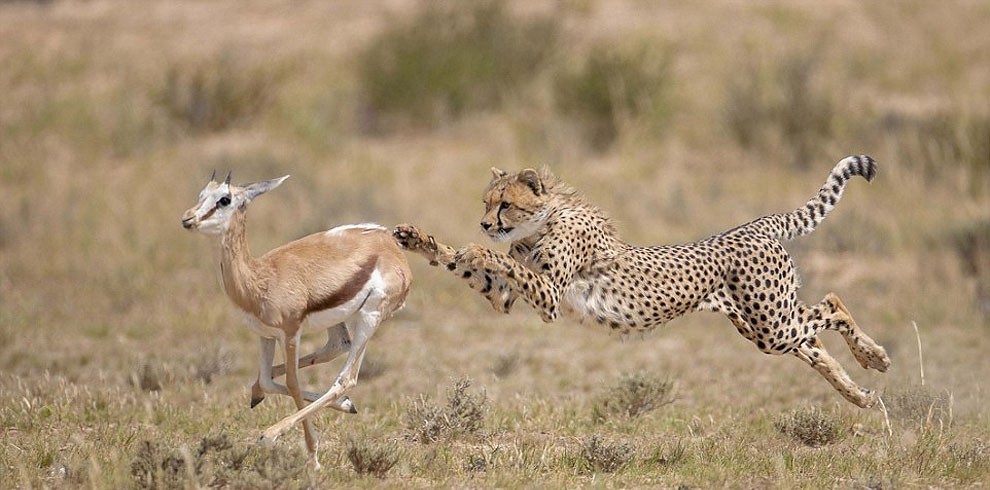} & 0 & 1 \\
        \hline
    \end{tabular}
    \caption{Sample entries from the structured dataset.}
    \label{tab:sample_entries}
\end{table}

\chapter{Diagram Representations and Quantum Circuits}

This section offers a detailed walk-through of each dataset, showcasing the string diagram, image diagram, their integration, and the corresponding quantum circuit representations.

\begin{figure}[h!] 
    \centering 
    \includegraphics[width=1.0\textwidth]{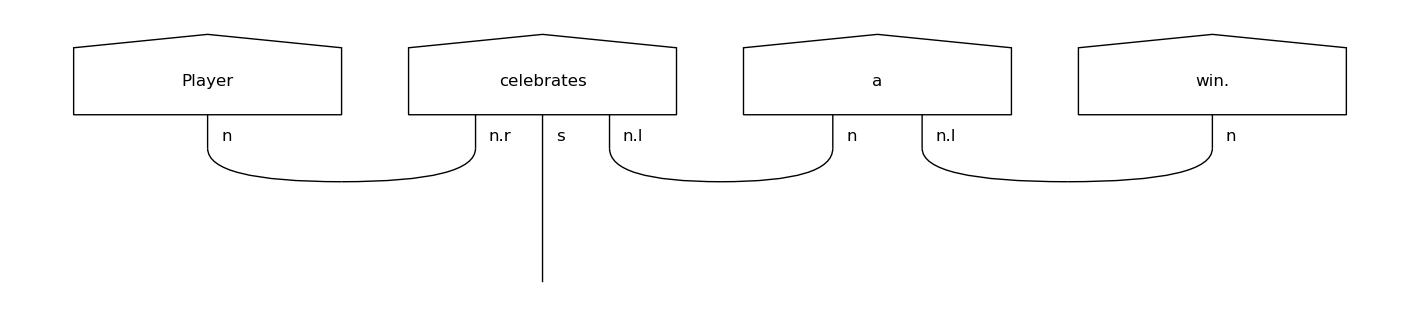} 
    \caption{Syntax-based string diagram representation.} 
    
\end{figure}

\begin{figure}
    \centering
    \includegraphics[width=1.0\linewidth]{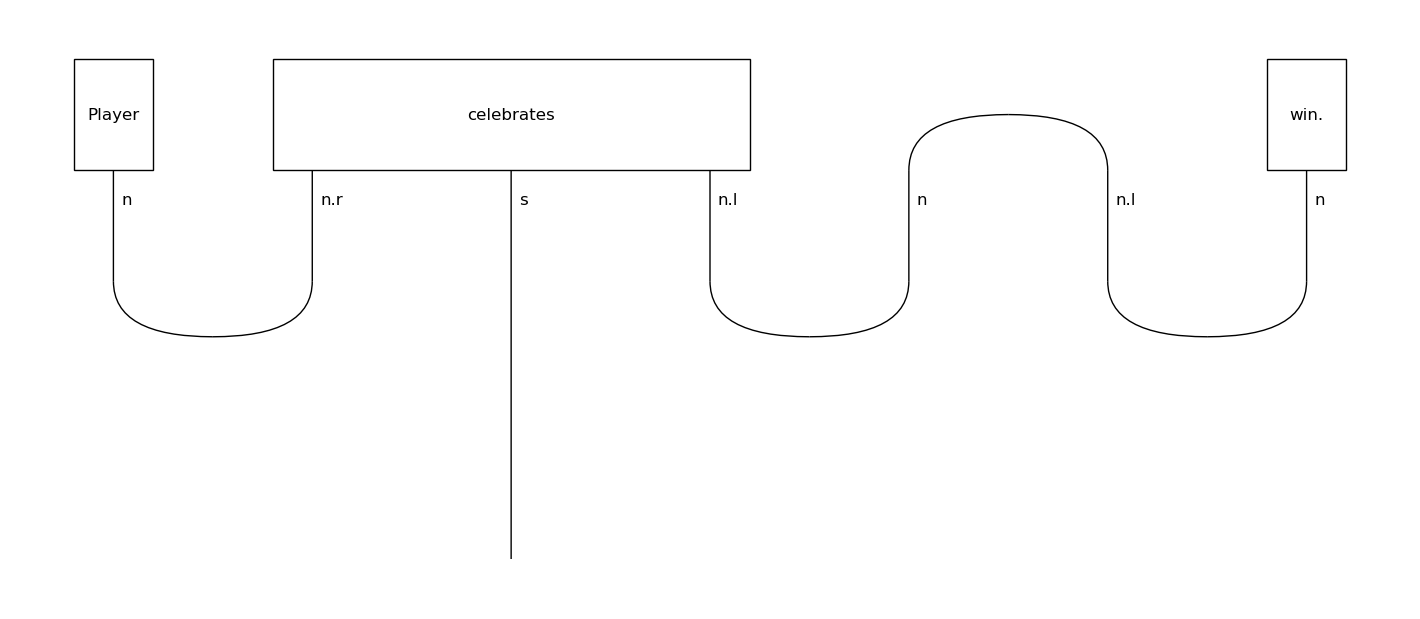}
    \caption{Syntax-based string diagram representation after removing cups.}
    
\end{figure}

\begin{figure}
    \centering
    \includegraphics[width=1.0\linewidth]{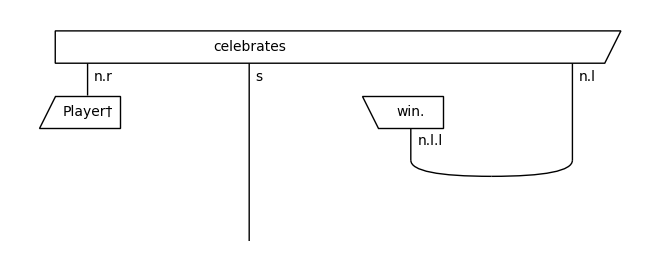}
    \caption{Syntax-based string diagram representation after rewrite function.}
    \label{fig:sentence_diag}
\end{figure}

\begin{figure}
    \centering
    \includegraphics[width=1\linewidth]{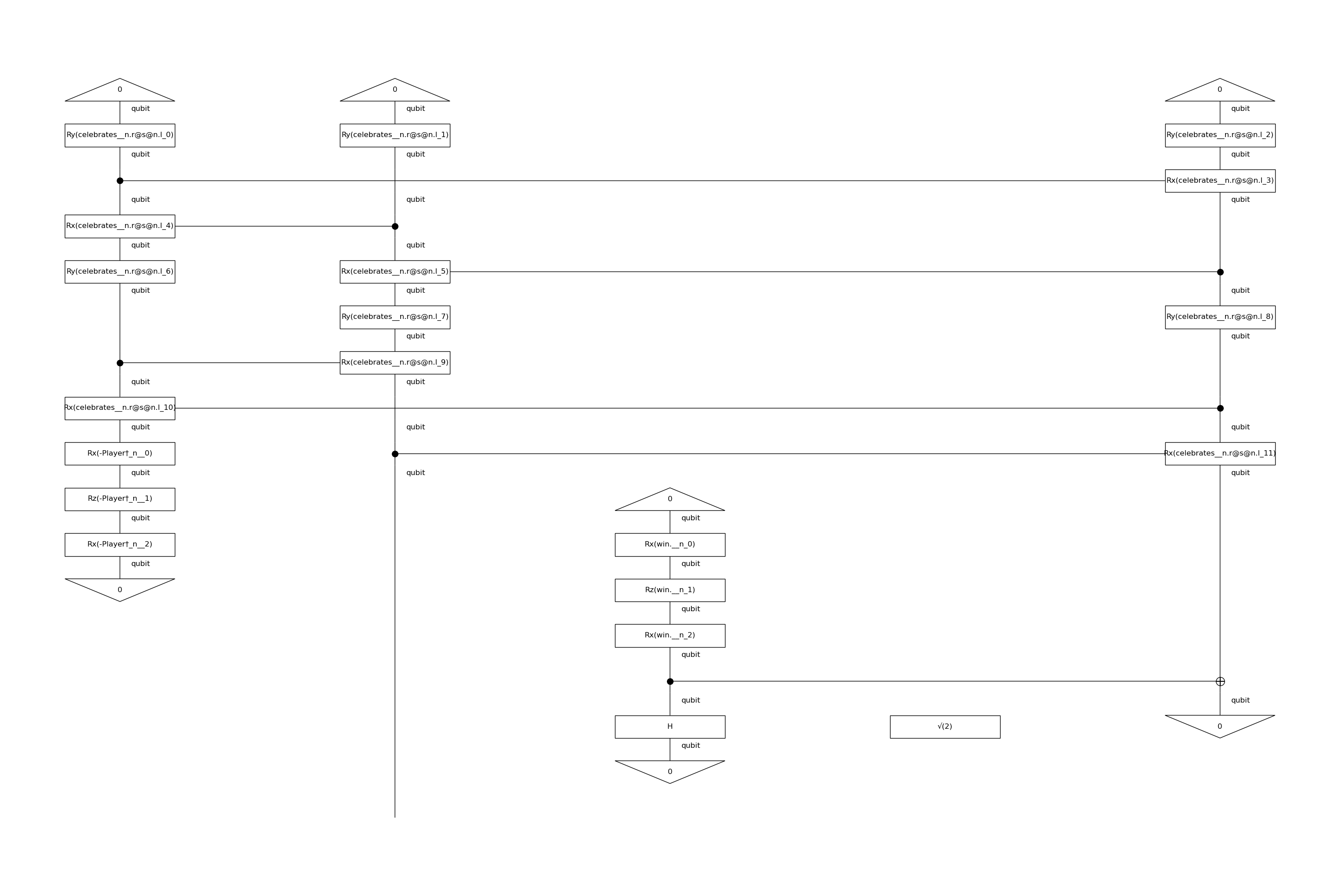}
    \caption{Quantum circuit representation for the sentence diagram in \ref{fig:sentence_diag}.}
    
\end{figure}

\begin{figure}
    \centering
    \includegraphics[width=1\linewidth]{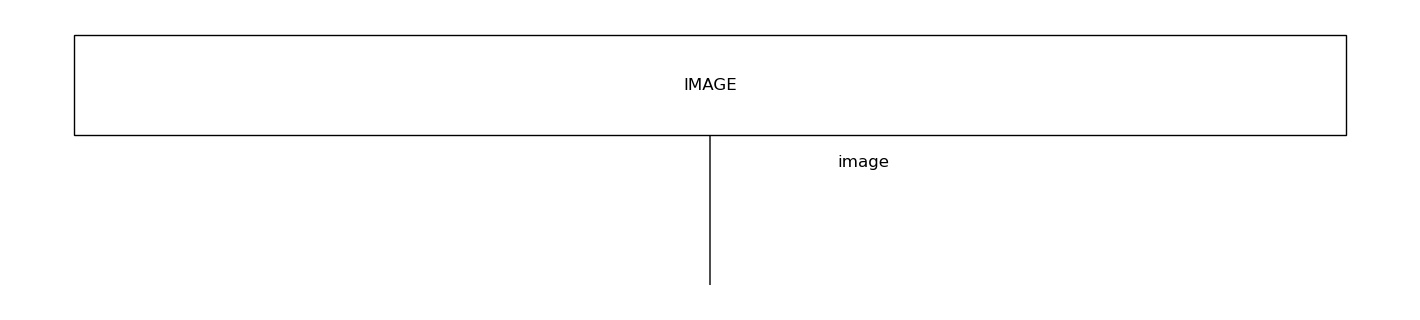}
    \caption{Image diagram representation.}
    \label{fig:image-diag}
\end{figure}

\begin{figure}
    \centering
    \includegraphics[width=1\linewidth]{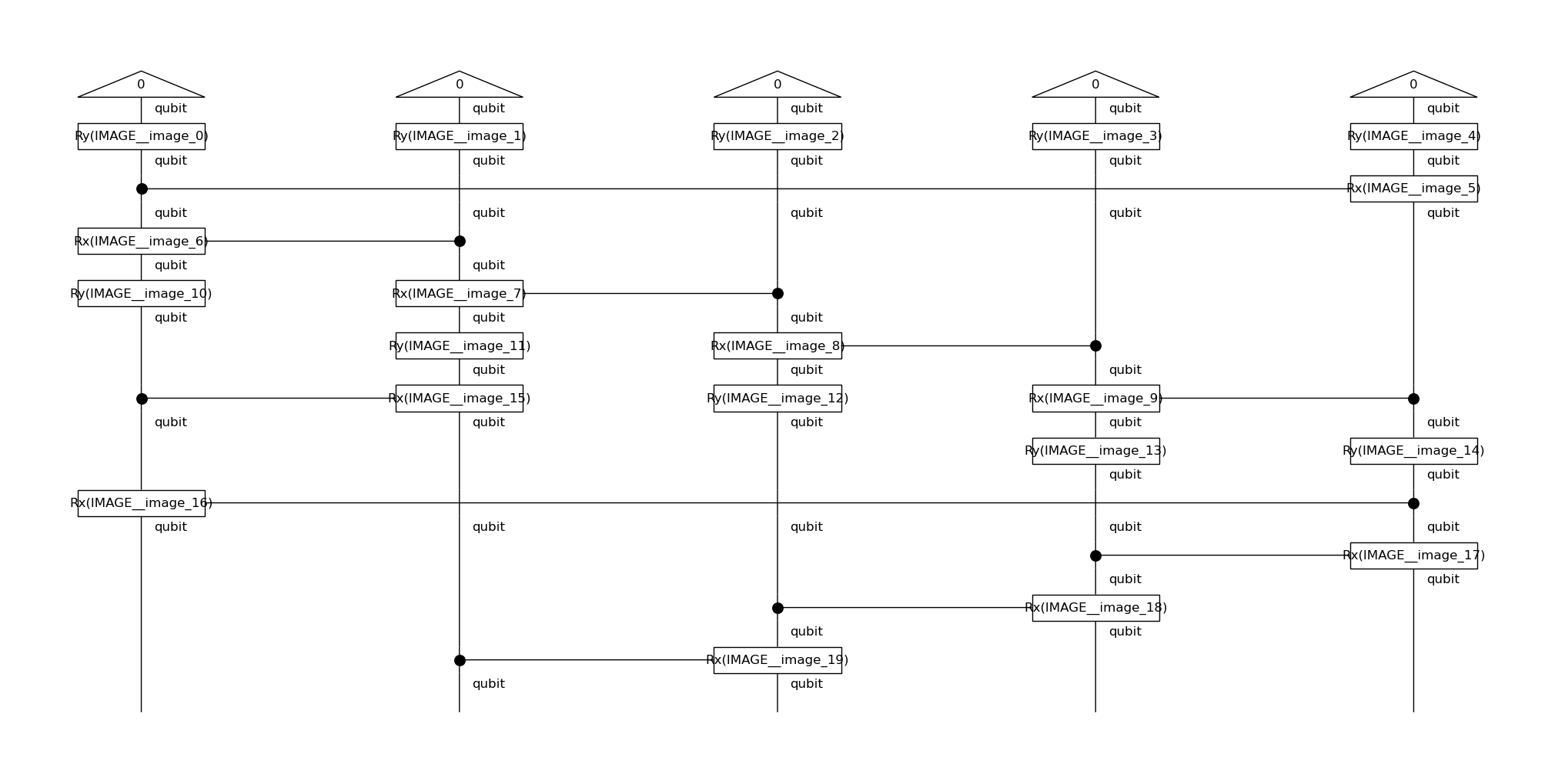}
    \caption{Quantum circuit representation for the image diagram in \ref{fig:image-diag}.}
    
\end{figure}

\begin{figure}
    \centering
    \includegraphics[width=1\linewidth]{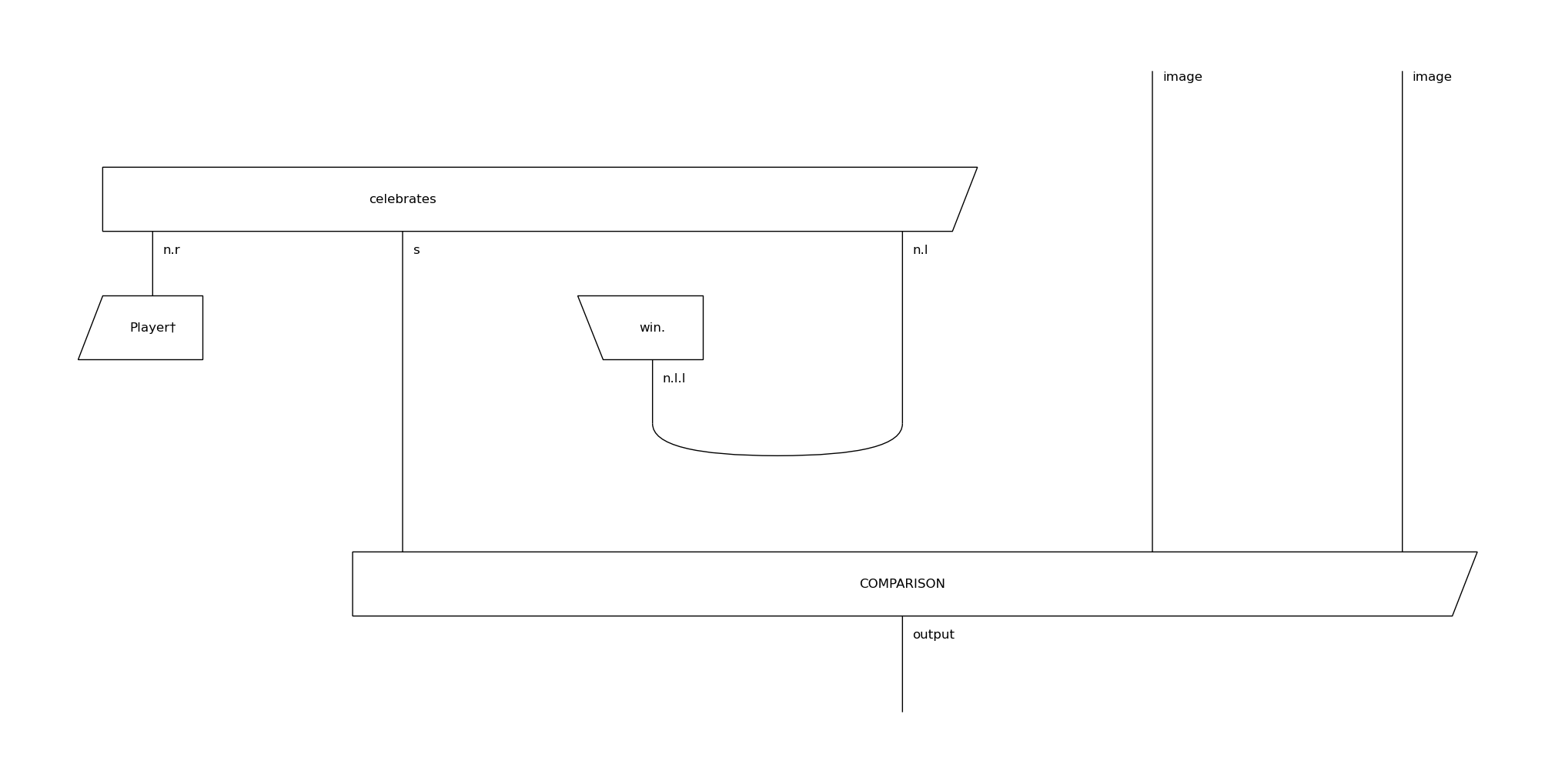}
    \caption{Concatenated diagram for unstructured dataset.}
    
\end{figure}

\begin{figure}
    \centering
    \includegraphics[width=1.0\linewidth]{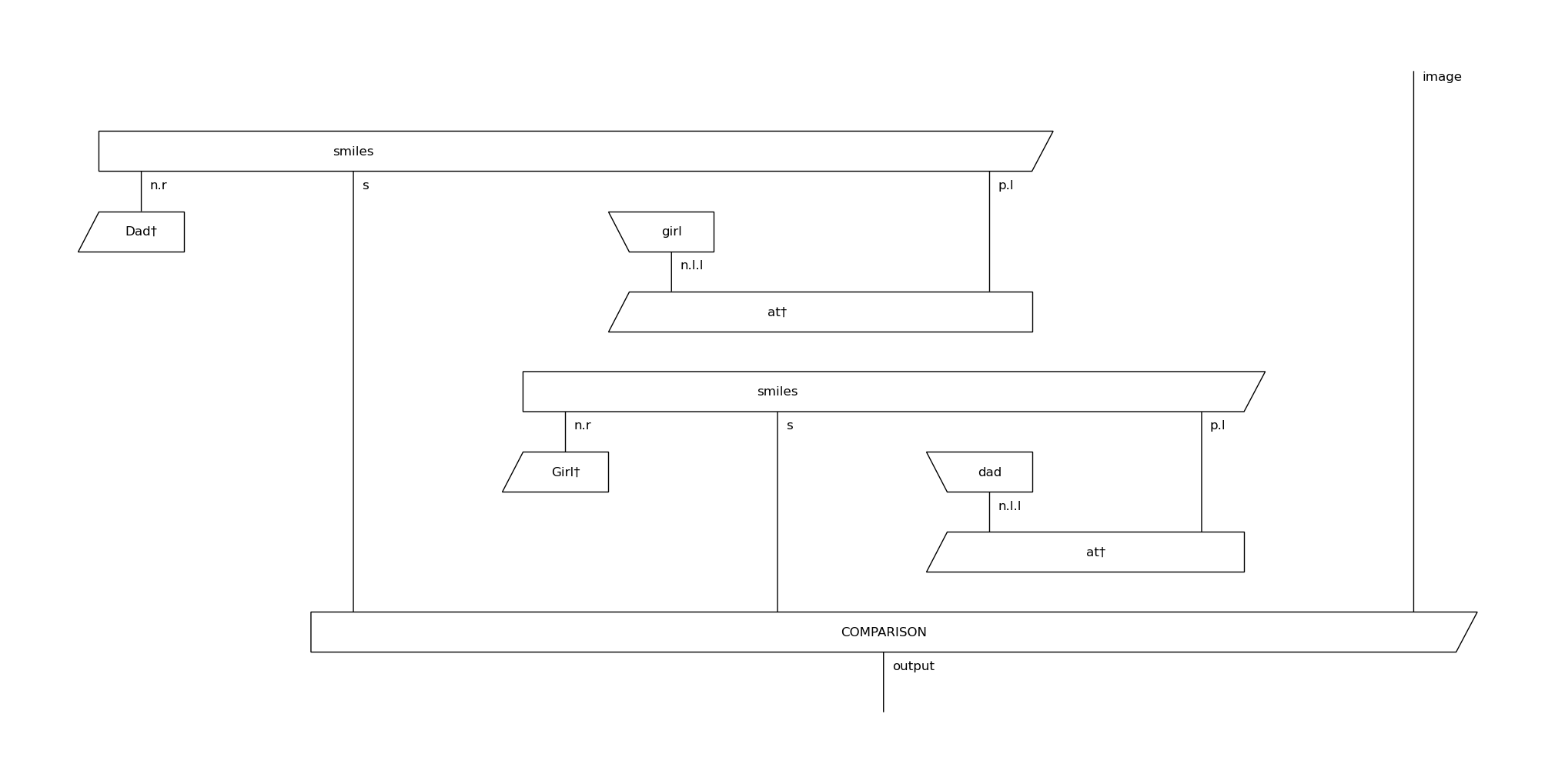}
    \caption{Concatenated diagram for structured dataset.}
    
\end{figure}

\begin{figure}
    \centering
    \includegraphics[width=1\linewidth]{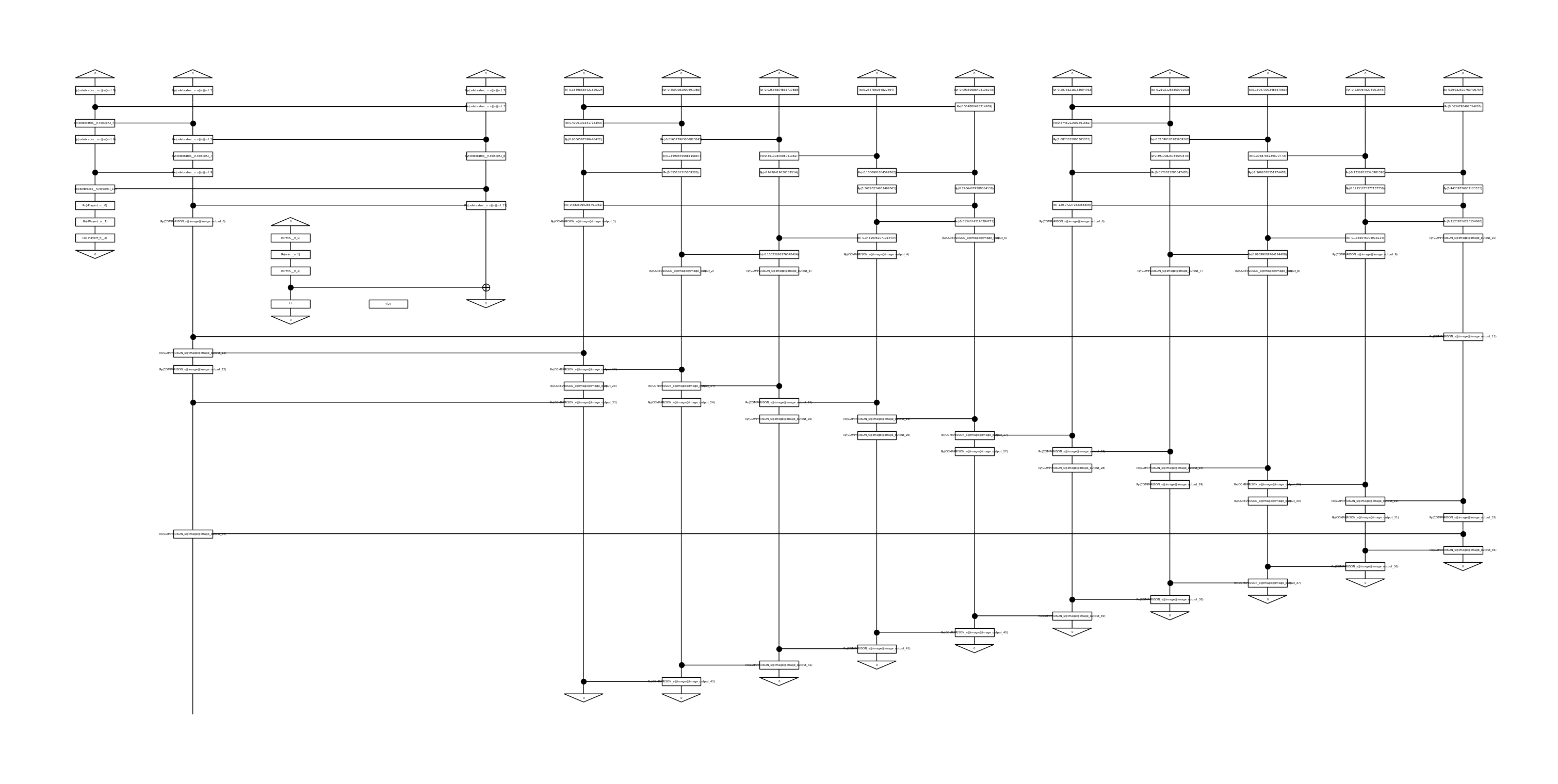}
    \caption{Concatenated quantum circuit for unstructured dataset.}
\end{figure}

\begin{figure}
    \centering
    \includegraphics[width=1\linewidth]{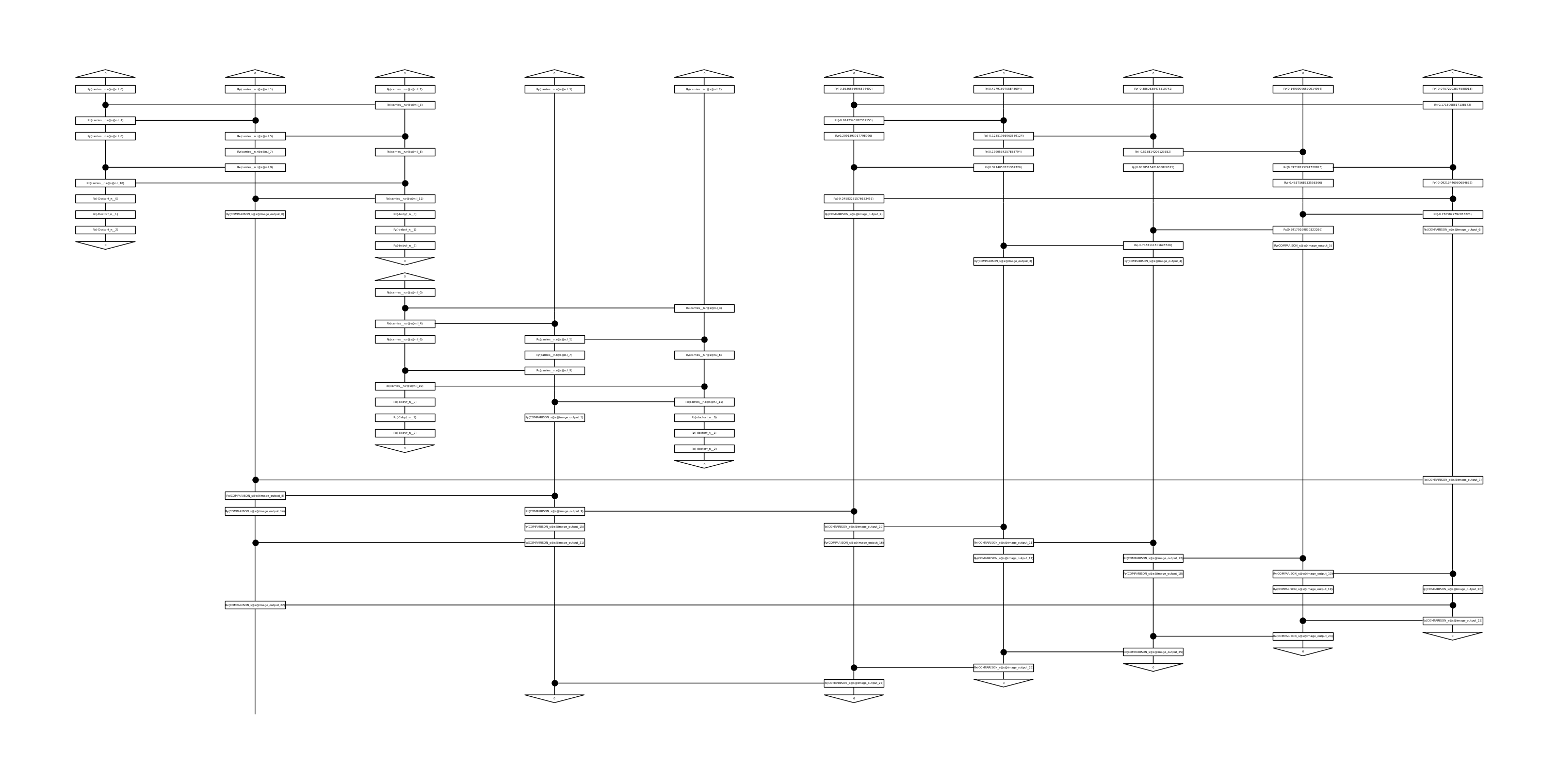}
    \caption{Concatenated quantum circuit for structured dataset.}
    
\end{figure}

\chapter{Results of Each Iteration}

\begin{table}[h!]
    \centering
    \begin{tabular}{|c|c|c|c|c|c|}
        \hline
        & \textbf{BoW} & \textbf{Cups} & \textbf{DisCoCat} & \textbf{Stairs} & \textbf{Tree} \\
        \hline
        1 & \textbf{73.58} & 56.6 & 63.64 & \textbf{79.25} & 66.04 \\
        \hline
        2 & 58.49 & 56.6 & 59.09 & 54.72 & \textbf{64.15} \\
        \hline
        3 & 67.92 & 56.6 & 56.82 & 47.17 & 54.72 \\
        \hline
        4 & 56.6 & 54.72 & \textbf{70.45} & 45.28 & 58.49 \\
        \hline
        5 & 52.83 & \textbf{58.49} & 65.91 & 32.08 & 60.38 \\
        \hline
        \rowcolor{lightgray2}\textbf{Average} & 61.884 & 56.602 & 63.182 & 51.7 & 60.756 \\
        \hline
    \end{tabular}
    \caption{Performance comparison for different compositional models for the unstructured dataset.}
    \label{tab:dataset_comparison}
\end{table}

\begin{table}[h!]
    \centering
    \begin{tabular}{|c|c|c|c|c|c|}
        \hline
        & \textbf{BoW} & \textbf{Cups} & \textbf{DisCoCat} & \textbf{Stairs} & \textbf{Tree} \\
        \hline
        1 & 50 & 50 & \textbf{68.75} & 45 & 55 \\
        \hline
        2 & 50 & 45 & 50 & 55 & 55 \\
        \hline
        3 & 50 & \textbf{60} & 56.25 & \textbf{65} & 55 \\
        \hline
        4 & 50 & 50 & 50 & 50 & 55 \\
        \hline
        5 & \textbf{50} & 40 & 50 & 55 & \textbf{60} \\
        \hline
        \rowcolor{lightgray2} \textbf{Average} & 50 & 49 & 55 & 54 & 56 \\
        \hline
    \end{tabular}
    \caption{Performance comparison for different compositional models for the structured dataset.}
    \label{tab:dataset_comparison}
\end{table}

\chapter{Code}
For additional details, code, and datasets related to this thesis, please visit the project repository on GitHub: \url{https://github.com/halaa901/QNLP-Thesis.git}.

\end{appendices}

\end{document}